\pdfoutput=1
\PassOptionsToPackage{table}{xcolor}
\documentclass{article} 
\usepackage[T1]{fontenc}
\usepackage{iclr2027_conference,times}

\usepackage{amsmath,amsfonts,bm}

\def\eqref#1{equation~\ref{#1}}

\def\1{\bm{1}}

\DeclareMathAlphabet{\mathsfit}{\encodingdefault}{\sfdefault}{m}{sl}
\SetMathAlphabet{\mathsfit}{bold}{\encodingdefault}{\sfdefault}{bx}{n}

\usepackage{hyperref}
\hypersetup{hidelinks}
\usepackage{url}
\usepackage{booktabs}       
\usepackage{multirow}
\usepackage{graphicx}
\usepackage{makecell}
\usepackage{amsmath}
\usepackage{amssymb}
\usepackage{subcaption}
\usepackage{siunitx}
\usepackage{pifont}
\usepackage{tabularx}
\usepackage{xcolor}
\usepackage{wrapfig}
\usepackage{nicematrix}

\usepackage{listings}
\usepackage{xcolor}
\usepackage{enumitem}

\usepackage{threeparttable} 
\usepackage{float}

\definecolor{codegreen}{rgb}{0,0.6,0}
\definecolor{codegray}{rgb}{0.5,0.5,0.5}
\definecolor{codepurple}{rgb}{0.58,0,0.82}
\definecolor{backcolour}{rgb}{0.95,0.95,0.92}

\definecolor{darkgreen}{rgb}{0.0, 0.8, 0.0}
\definecolor{darkred}{rgb}{0.8, 0.0, 0.0}

\lstdefinestyle{mystyle}{
    backgroundcolor=\color{backcolour},   
    commentstyle=\color{codegreen},
    keywordstyle=\color{magenta},
    numberstyle=\tiny\color{codegray},
    stringstyle=\color{codepurple},
    basicstyle=\ttfamily\footnotesize,
    breakatwhitespace=false,         
    breaklines=true,                 
    captionpos=b,                    
    keepspaces=true,                 
    numbers=left,                    
    numbersep=5pt,                  
    showspaces=false,                
    showstringspaces=false,
    showtabs=false,                  
    tabsize=2,
    frame=single,
    framerule=0pt,
    rulecolor=\color{black},
}
\title{InfoVis: Unified Slide Image Generation and Editing}

\newcommand{\coderepo}{\url{https://github.com/VectorSpaceLab/InfoVis}}

\author{\textbf{Jiahao Wang$^{1,2,\ast}$ \quad Xiyan Jiang$^{2,3,\ast}$ \quad Chenyuan Wu$^{4,\ast}$ \quad Haoqiang Zhang$^{4,\dagger}$ \quad Xin Luo$^{4}$} \\
\textbf{Yueze Wang$^{2}$ \quad Defu Lian$^{4}$ \quad Bo Zhang$^{3}$ \quad Jiajun Zhang$^{1}$ \quad Zheng Liu$^{2,\ddagger}$} \\
$^{1}$Institute of Automation, Chinese Academy of Sciences \\
$^{2}$Beijing Academy of Artificial Intelligence \\
$^{3}$Zhejiang University \\
$^{4}$University of Science and Technology of China \\
Code: \coderepo
}

\iclrfinalcopy
\begin{document}

\maketitle
\lhead{Preprint}
{\renewcommand\thefootnote{}%
\footnotetext{$^{\ast}$Equal contribution. \quad $^{\dagger}$Core contributor. \quad $^{\ddagger}$Corresponding author.}}

\begin{abstract}
  
    Generating and editing presentation slides requires models to combine accurate
    text rendering, instruction following, visual grounding, aesthetic design, and
    precise editing within one canvas. Existing work covers only subsets of these
    capabilities and lacks large-scale slide-native supervision and diagnostic
    evaluation. We formulate four slide-centric tasks and build a scalable data
    engine with over 13M continual-pretraining records and 1.9M supervised
    fine-tuning records. We also introduce VisSlideBench, a benchmark of 1,527
    manually screened instances that covers all four tasks with task-specific,
    multidimensional evaluation. Its automatic rankings agree with blinded,
    dimension-wise human rankings. Using these resources, we train InfoVis, a
    unified 9B slide generation and editing model. On VisSlideBench, InfoVis leads
    every evaluated open-source model on all eleven metrics and scores 7.94
    overall against 7.99 for Nano-Banana-2. Starting from
    FLUX.2-klein-base-9B, slide-specific post-training adds 1.6 points
    overall, with the largest generation gains in content faithfulness; the remaining
    gap to proprietary models is smallest in content faithfulness and largest
    in aesthetic quality. Built on InfoVis, our
    document-to-slide workflow achieves the second-highest Aesthetics score on
    the external SlidesGen-Bench. We will release the model, code, data, and
    benchmark.

    \end{abstract}
        
\section{Introduction}
\label{sec:intro}

Image generation has rapidly expanded beyond natural-image synthesis toward structured, text-rich visual content, including posters, advertisements, infographics, and presentation slides~\citep{wang2025designdiffusion,peng2025bizgen,gao2025postermaker,chen2025postercraft,hu2025dreamposter}. Slides are particularly demanding because capabilities often developed separately must work together reliably within a single canvas. Models must render text accurately, arrange heterogeneous elements into coherent layouts, incorporate supplied visual assets faithfully, and apply local or global edits while preserving unrelated content. Recent open-source image generators have made substantial progress~\citep{flux-2-2025,team2025zimage,wu2025qwenimage,LongCat-Image}, yet they still combine these capabilities unreliably. Accurate, polished, and controllable slide generation and editing therefore remain difficult.

A major obstacle is the lack of slide-native supervision and diagnostic evaluation. General-purpose image generators are trained on broad visual distributions, with limited supervision for the content--layout--design relationships characteristic of presentation slides. Meanwhile, existing benchmarks primarily target programmatic or deck-level presentation generation, or include slides only as a subset of broader visual-content tasks~\citep{ge2025autopresent,huang2025pptbench,li2026bizgeneval,chen2026presentbench,yang2026slidesgen}. These settings do not systematically isolate the capabilities required for single-slide image generation and editing.

\begin{figure}[t]
    \centering
    \includegraphics[width=\linewidth]{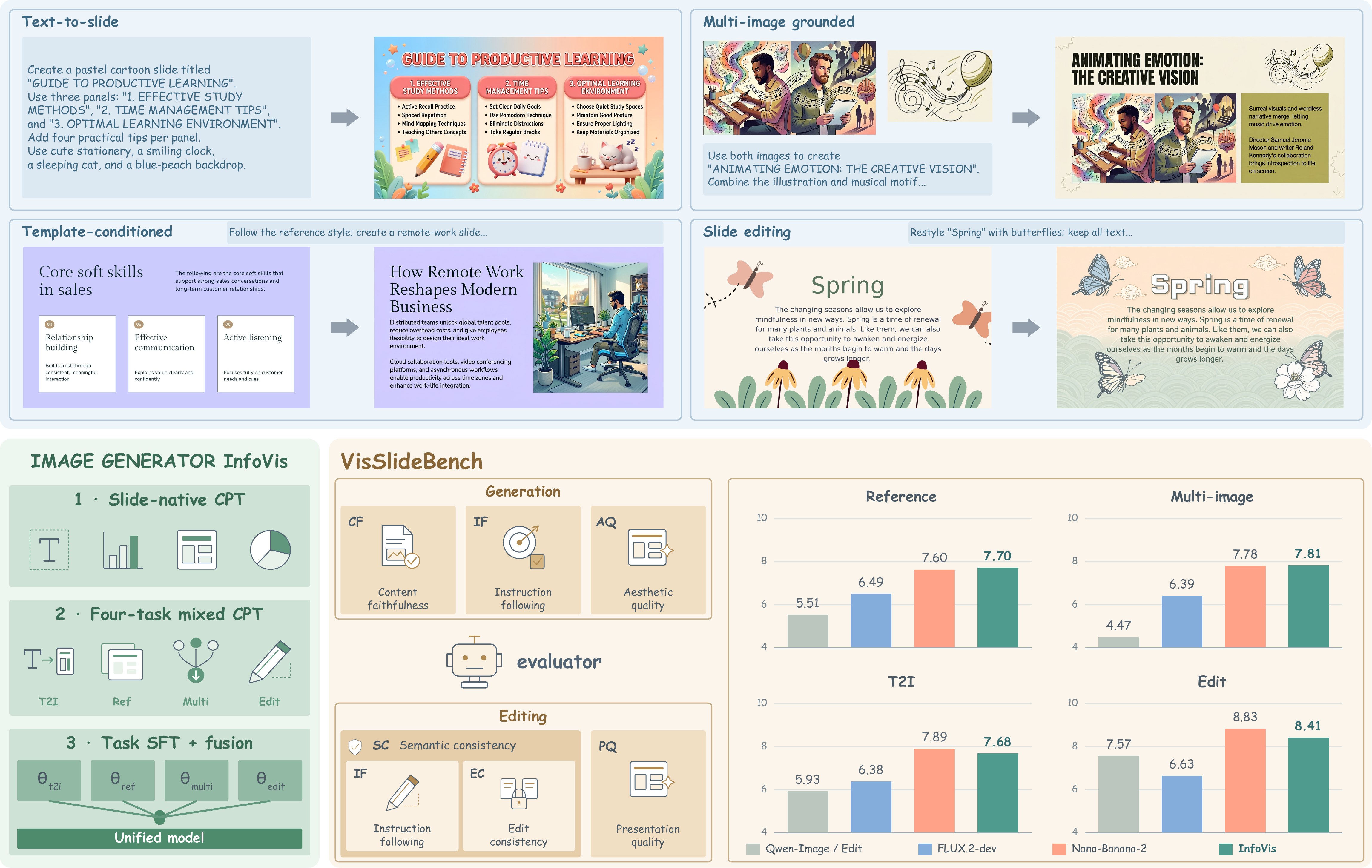}
    \caption{Overview of our slide tasks, InfoVis, and VisSlideBench.
    InfoVis is trained through three stages on four generation and editing
    tasks, and VisSlideBench evaluates each task along task-specific
    dimensions. Bar values are the dimension means of
    Tables~\ref{tab:t2i_fine_grained_results}--\ref{tab:edit_fine_grained_results}.}
    \label{fig:overview}
\end{figure}

In this work, we cast slide creation as a unified conditional visual synthesis problem and study four tasks under a shared text-and-image interface: text-to-slide generation, template-conditioned generation, multi-image grounded generation, and instruction-based slide editing (Section~\ref{sec:task_formulation}). The tasks share requirements for accurate content rendering and coherent visual composition, while visual inputs serve as style references, content assets, or source slides to edit, which motivates supporting all four within a unified model. Figure~\ref{fig:overview} summarizes our data, model, and benchmark.

This shared view directly shapes our data design. We build a scalable \textbf{slide-native data engine} that combines real presentations with synthetic supervision and covers the conditioning requirements of all four tasks. Its instructions range from concise user requests to detailed descriptions and structured \texttt{style\_json} specifications, so that one model serves both human-facing and machine-generated design conditions. Editing pairs come from deterministic PPTX transformations with instructions derived from the executed changes, complemented by model-generated targets for open-ended visual transformations. The resulting resource contains over 13M continual-pretraining records and 1.9M supervised fine-tuning records.

To systematically diagnose slide generation and editing capabilities, we introduce \textbf{VisSlideBench}, a benchmark of 1,527 manually screened instances dedicated to single-slide image generation and editing across all four tasks. VisSlideBench measures complementary dimensions of slide quality. It evaluates generation in terms of Content Faithfulness, Instruction Following, and Aesthetic Quality, and editing in terms of Semantic Consistency and Presentation Quality. The generation track spans the same three prompt settings as the training data. The automatic evaluator's dimension-wise rankings agree with blinded human judgments.

Using these resources, we train \textbf{InfoVis}, a unified 9B slide generation
and editing model based on FLUX.2-klein-base-9B~\citep{flux2klein2026}.
On VisSlideBench, InfoVis reaches an overall score of 7.94, leading every
evaluated open-source model on all eleven metrics, against 7.99 for
Nano-Banana-2~\citep{nanobanana2}. Our analysis yields three findings:
the gains come from slide-specific supervision rather than scale, as InfoVis
is trained from FLUX.2-klein-base-9B and outperforms the 32B
FLUX.2-dev; unification costs little relative to our own task specialists,
unlike existing unified open-source models, which trail specialists by a wide
margin; and the remaining gap to proprietary models is ordered by
subjectivity, from content faithfulness through instruction following to
aesthetic quality.
Beyond VisSlideBench, InfoVis is the best open-source model in the BizGenEval
Slides comparison~\citep{li2026bizgeneval}, and our document-to-slide workflow
ranks second among ten systems in Aesthetics on
SlidesGen-Bench~\citep{yang2026slidesgen}.

Our main contributions are:
\begin{itemize}[leftmargin=*]
    \item We formulate four slide generation and editing tasks and build a \textbf{slide-native data engine} with over 13M continual-pretraining and 1.9M supervised fine-tuning records.

    \item We introduce \textbf{VisSlideBench}, a benchmark for single-slide image generation and editing that covers all four tasks with task-specific, multidimensional evaluation.

    \item We develop \textbf{InfoVis}, a unified 9B model that leads every evaluated open-source model on all eleven VisSlideBench metrics, scores 7.94 overall against 7.99 for Nano-Banana-2, and is the best open-source model on BizGenEval Slides. Built on InfoVis, our document-to-slide workflow ranks second among ten systems in Aesthetics on SlidesGen-Bench.
\end{itemize}

\section{Related Work}
\label{sec:related}

\paragraph{Structured Rich-Text and Design Image Generation}

Visual text generation methods improve text fidelity through specialized
glyph, position, layout, and text-attribute representations
\citep{jian2023glyphdraw,chen2023textdiffuser,
chen2024textdiffuser-2,tuo2024anytext,tuo2024anytext2,
shuai2026glyphprinter,mao2026residual}.
Building on these advances, recent datasets and models extend text-aware
generation to dense and paragraph-level scenarios
\citep{wang2025textAtlas5M,wang2025longtextar}.
Beyond rendering text accurately, another line of work studies the
generation of complete design images, including posters, product
advertisements, infographics, and slides
\citep{wang2025designdiffusion,peng2025bizgen,
gao2025postermaker,chen2025postercraft,hu2025dreamposter,
cui2026simpleposter}.
Related work also studies factual consistency during structured visual
generation and editing~\citep{factualitymatters}, emphasizing that visual
quality must be coupled with faithful content realization. General-purpose
generators~\citep{flux-2-2025,team2025zimage,wu2025qwenimage} and unified
generation--editing models~\citep{deng2025bagel,wu2026omnigen2} handle diverse
inputs but lack slide-native supervision, while the methods above typically
optimize isolated visual artifacts and do not jointly address slide-specific
template conditioning, multi-image grounding, and instruction-based editing.
InfoVis unifies these capabilities in a slide-centric generator with shared
content, layout, and design supervision.

\paragraph{Benchmarks for Slide Generation}

Existing benchmarks evaluate slide generation from complementary
perspectives. SlidesBench~\citep{ge2025autopresent} focuses on
programmatic single-slide generation, while
EvoPresent~\citep{liu2025presenting} evaluates end-to-end
paper-to-presentation systems. PPTBench~\citep{huang2025pptbench} covers
PowerPoint understanding, modification, and generation, whereas
BizGenEval~\citep{li2026bizgeneval} treats slides as one category of
broader commercial visual content. More recent benchmarks, including
PresentBench~\citep{chen2026presentbench},
SlidesGen-Bench~\citep{yang2026slidesgen}, and
DECKBench~\citep{jang2026deckbench}, introduce fine-grained evaluation
protocols for slide generation: SlidesGen-Bench separates content, aesthetic,
and editability criteria for reproducibility, and DECKBench further extends
evaluation to multi-turn slide editing. These benchmarks nevertheless remain
centered on complete presentations and agentic workflows, and provide limited
diagnosis of individual slide capabilities. VisSlideBench instead targets
single-slide image generation and editing, evaluating text fidelity, template
consistency, visual grounding, and edit preservation with task-specific
protocols and metrics.
\section{Task Formulation and Slide Data Engine}
\label{sec:data}

\begin{figure}[t]
    \centering
    \includegraphics[width=\linewidth]{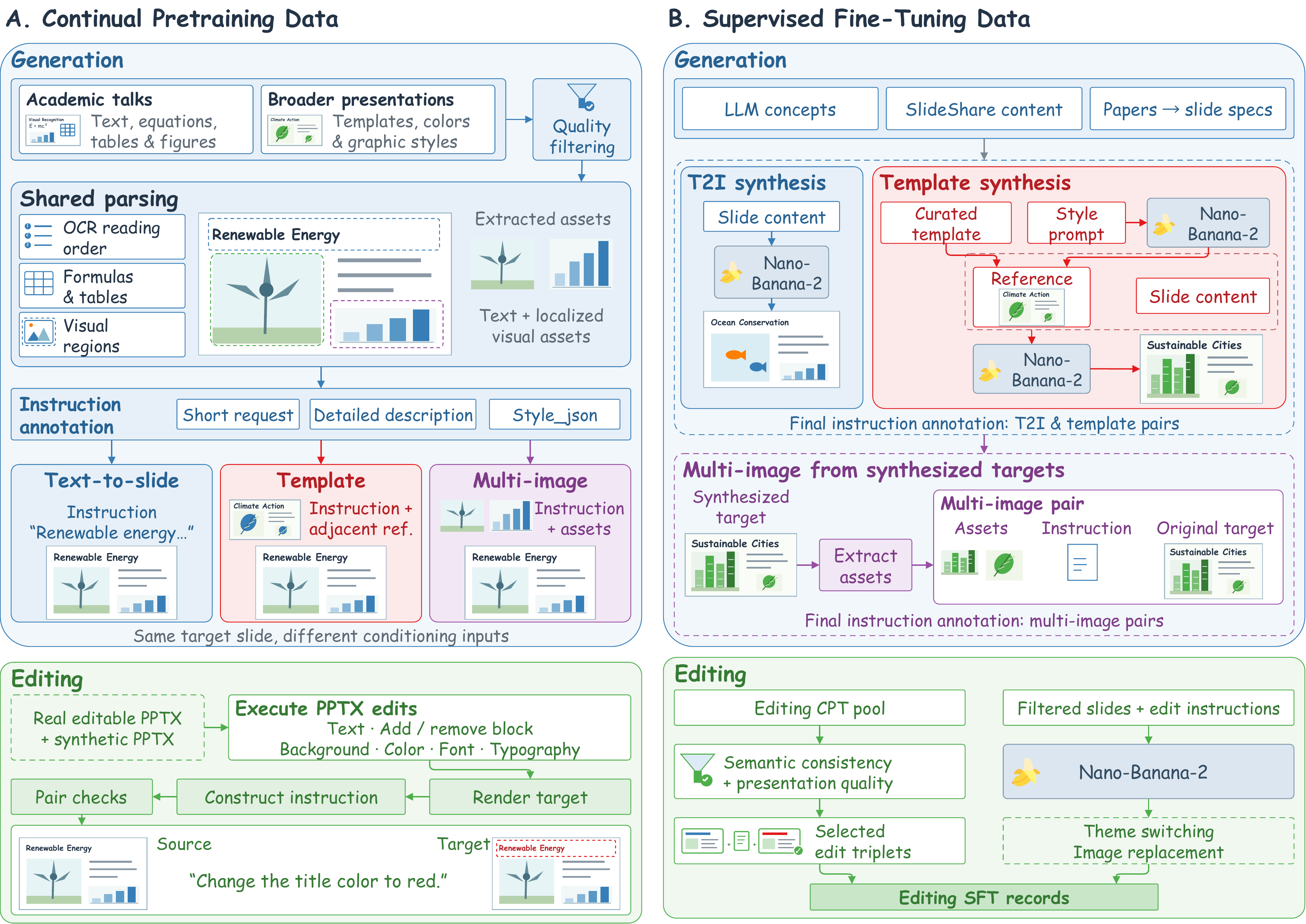}
    \caption{Slide data engine. (A) Continual pretraining: real decks are
    filtered, parsed into text and localized visual assets, and annotated
    with short requests, detailed descriptions, and \texttt{style\_json}; the
    same target slide yields text-to-slide, template-conditioned, and
    multi-image records, and executed PPTX edits yield
    source--instruction--target triplets. (B) Supervised fine-tuning:
    Nano-Banana-2 synthesizes targets from LLM concepts, SlideShare content,
    and paper-derived slide specifications; multi-image records reuse assets
    extracted from synthesized targets; editing records combine selected CPT
    triplets with model-generated theme switching and image replacement.}
    \label{fig:data_engine}
\end{figure}

Our slide data engine recovers slide content and visual elements from
presentation pages and converts them into task-specific conditions while
keeping the complete slide as the target; editing data instead comes from
controlled changes to editable slides. The four tasks thus share content,
layout, and design supervision. Appendix~\ref{app:data} gives statistics and
implementation details. Figure~\ref{fig:data_engine} summarizes the
construction.

\subsection{Task Formulation}
\label{sec:task_formulation}

We formulate slide creation as conditional visual synthesis
$y=G_\theta(x,I)$ over a textual condition $x$ and optional input images $I$.
In \emph{text-to-slide generation}, $I=\varnothing$; in
\emph{template-conditioned generation}, $I$ is a reference slide that
specifies the visual style; in \emph{multi-image grounded generation}, $I$
holds one or more visual assets that must be preserved in the composition; in
\emph{slide editing}, $I$ is the source slide and $x$ the requested change,
with unrelated content kept stable.

Generation instructions take three forms: a short caption, a detailed
description of content and visual design, and a structured
\texttt{style\_json} of the page composition; editing uses a natural-language
instruction. The same interfaces serve data construction and evaluation, so
the training tasks align with VisSlideBench.

\subsection{Continual Pretraining Data}
\label{sec:data_cpt}

\noindent\textbf{Real-slide sources and parsing.}
We collect real presentations from two complementary sources: author-released
conference talks with dense text, equations, and scientific figures, and a
broader collection spanning 39 topical categories with more diverse templates
and styles. We use Qwen3.5-27B~\citep{qwen3.5} to remove low-quality
and non-slide content, then parse the retained pages with
PaddleOCR-VL~\citep{cui2025paddleocrvl} into text in reading order and
localized visual elements, which form the shared content representation for
the generation tasks.

\noindent\textbf{Generation task construction.}
From each parsed slide, we construct three generation views that share the
original rendered slide as the target while using different conditioning
inputs. An open VLM generates a short request, a detailed description, and a
structured \texttt{style\_json} specification from the slide, its OCR
transcription, and extracted visual elements. Text-to-slide records use the textual condition alone; the instruction therefore describes the whole slide, visual assets included.
Template-conditioned records pair the target with an adjacent slide from the
same deck as the reference; we retain pairs with consistent backgrounds,
colors, typography, and decorative styles. Multi-image grounded records reuse
one to four localized visual elements as conditioning inputs and describe how
they should be incorporated into the slide.

\noindent\textbf{Executable editing pairs.}
\label{sec:data_edit}
For editing CPT, we construct source--instruction--target triplets through
executable changes to PPTX files drawn from real editable presentations and
synthetic PPTX files built from OCR text and presentation backgrounds. Seven
operations cover text modification, block addition and removal, background
updates, and changes to color, font, and typography; each is executed in the
PPTX file and rendered, and an LLM describes the realized change from the
source image, target image, and executed action. We remove source slides with
layout or rendering defects and pairs with failed edits or inconsistent
instructions. Across the four task branches, continual pretraining contains
over 13M records.

\subsection{Supervised Fine-Tuning Data}
\label{sec:data_sft}

\noindent\textbf{Generation SFT.}
Generation SFT content comes from three sources: LLM-generated concepts
across diverse domains, structured content from real SlideShare decks, and
research papers converted into slide-level specifications. Nano-Banana-2
renders the content into target slides under diverse visual styles; for
template-conditioned generation, we first obtain a reference slide from a
curated template library or by asking Nano-Banana-2 to create one, then
generate the target from the reference and new content. For multi-image
grounded generation, we extract charts, tables, and other visual regions from
the synthesized targets as conditioning inputs. Qwen3.5-27B annotates the
records with short requests, detailed descriptions, and structured
\texttt{style\_json} specifications, or with instructions describing how the
assets should be incorporated, and we filter invalid inputs and inconsistent
instructions.

\noindent\textbf{Editing SFT.}
For the seven executable operations, we select high-quality triplets from
the editing CPT pool by semantic consistency and presentation quality, and add
theme switching and image replacement examples generated by Nano-Banana-2
from filtered real slides, recaptioned so that each instruction describes the
realized change. Across the four task branches, the SFT resource contains
about 1.9M records (Appendix~\ref{app:data_sft}).

\section{VisSlideBench}
\label{sec:bench}

\subsection{Benchmark Task Suite}
\label{sec:benchmark_task_suite}

\textbf{VisSlideBench} (Figure~\ref{fig:visslidebench} in Appendix~\ref{app:benchmark}) evaluates the generation and editing of individual
slide images across the four tasks defined in Section~\ref{sec:task_formulation}.
It contains 1,527 manually screened instances: 943 in
\emph{VisSlideBench-Gen}, covering text-to-slide, template-conditioned, and
multi-image grounded generation, and 584 in \emph{VisSlideBench-Edit}.
Generation prompts range from short requests to detailed descriptions and
structured \texttt{style\_json} specifications, covering instructions from
users and agents.

Among the benchmarks compared in Appendix~\ref{app:benchmark_comparison},
VisSlideBench is the only one that jointly covers all four creation
interfaces, three prompt granularities, and task-specific measures of content
fidelity, instruction compliance, aesthetics, and visual grounding.

\subsection{Benchmark Construction}
\label{sec:benchmark_construction}

\paragraph{Generation.}
\label{sec:generation_benchmark_construction}
VisSlideBench-Gen combines 493 cases derived from real presentations with
450 cases whose target slides are generated by
Nano-Banana-2~\citep{nanobanana2}. The real cases preserve authored content and
design patterns, while the synthetic cases broaden visual-design coverage.
Following Section~\ref{sec:data}, we construct the instructions and visual
inputs required for each case: no image for text-to-slide, one reference slide
for template-conditioned generation, and one to four assets for multi-image
grounded generation. We manually screen target slides and instructions,
removing slides of poor visual quality and prompts that are ambiguous or
inconsistent.

\paragraph{Editing.}
\label{sec:editing_benchmark_construction}
VisSlideBench-Edit uses a subset of the editing data reserved for evaluation
(Section~\ref{sec:data_edit}); because these cases come from the same
pipeline as the editing training data, the split measures in-distribution
editing quality (Appendix~\ref{app:limitations}). Each instance contains a
source slide, an edit instruction, and the corresponding target. We remove ambiguous instructions,
invalid pairs, imperceptible edits, and severe visual artifacts through manual
screening. The nine operations cover text modification, block addition and
removal, background updates, changes to color, font, and typography, image
replacement, and theme switching. Appendix~\ref{app:benchmark_composition}
reports task counts, source composition, domains, and editing operations,
and describes how all cases are held out from training.

\subsection{Evaluation Protocol}
\label{sec:evaluation_protocol}

We use OCR to evaluate text rendering and GPT-4.1~\citep{openai2025gpt41} to
judge visual content, compliance with design instructions, and presentation
quality. The criteria are evaluated separately before task-level aggregation;
the text-scoring formula and additional evaluation details appear in
Appendix~\ref{app:benchmark_metrics}.

\paragraph{Content Faithfulness (CF).}
\label{sec:generation_evaluation}
We score text rendering with the normalized edit distance of
\citet{gong2025seedream20nativechineseenglish}, adapted to the text the
instruction requires (Appendix~\ref{app:benchmark_metrics}). For multi-image
grounded generation, CF additionally checks whether every required input image
is faithfully preserved and appropriately incorporated into the generated
slide.

\paragraph{Instruction Following (IF) and Aesthetic Quality (AQ).}
IF measures whether the output realizes the explicit design requirements in
the instruction, judged along Background Compliance, Color Palette Adherence,
Typography Compliance, Layout and Position Accuracy, Mood and Atmosphere
Match, and Block-Level Style Fidelity. AQ measures the intrinsic visual quality
of the slide and is judged from the output image alone, without the
instruction, along Clarity, Layout, Style Coherence, and Presentation
Professionalism. The
separation matters because a slide can satisfy every design constraint yet
remain visually weak, or look attractive while violating the instruction.
Appendix~\ref{app:benchmark_metrics} defines each dimension.

\paragraph{Editing.}
\label{sec:editing_evaluation}
Adapting the SC/PQ framework of VIEScore~\citep{ku2024viescore} to slide
editing, the evaluator receives the source
slide, editing instruction, and model output to score Instruction Following
($S_{\mathrm{IF}}$), whether the requested change is applied, and Edit
Consistency ($S_{\mathrm{EC}}$), preservation of unrelated text, images,
layout, and styles. We define Semantic Consistency as
$S_{\mathrm{SC}}=\min(S_{\mathrm{IF}},S_{\mathrm{EC}})$, so a high score
requires both. Presentation Quality ($S_{\mathrm{PQ}}$) uses the same four
dimensions as generation AQ and is likewise judged from the output image
alone. The overall editing score is
$S_{\mathrm{Edit}}=\sqrt{S_{\mathrm{SC}}S_{\mathrm{PQ}}}$.

\subsection{Human Validation}
\label{sec:human_validation}

We assess agreement with human judgments using the annotation protocol of
EditScore~\citep{luo2026editscore}: two groups of two domain experts each
produce a consensus ranking of five anonymized model outputs per example,
separately for each evaluation dimension, across all four tasks, yielding
2,153 valid pairwise comparisons.
GPT-4.1 reaches a mean Kendall's $\tau_b$ of $0.640$, a mean Spearman's
$\rho$ of $0.700$, and pairwise agreement of $88.71\%$.
Appendix~\ref{app:benchmark_human} gives the protocol and per-task results.

\section{InfoVis Training}
\label{sec:method}

\paragraph{Slide-native model training.}
We adapt the pretrained image generator FLUX.2-klein-base-9B~\citep{flux2klein2026} to slide images, keeping its architecture and raising the maximum prompt length to 4,096 tokens. Training proceeds in three stages.
\textbf{Stage 1: Slide-native continual pretraining.}
We first continue pretraining the base model on large-scale text-to-slide data with short captions, long descriptions, and structured design prompts, adapting it to the slide domain and learning shared priors for jointly coordinating text rendering, layout, visual hierarchy, and design.
\textbf{Stage 2: Four-task mixed continual pretraining.}
We then train on a mixture of the four task datasets, extending these priors to style-reference following, multi-image grounding, and image editing.
\textbf{Stage 3: Task-specific supervised fine-tuning and model fusion.}
Finally, we perform one epoch of text-to-slide SFT to reduce the generation--editing distribution gap, fine-tune four task-specific models from the resulting shared checkpoint $\theta_{\mathrm{shared}}$, and linearly fuse them into the final model, following weight-space merging of fine-tuned checkpoints~\citep{wortsman2022modelsoups,ilharco2023taskarithmetic}.

\paragraph{Document-to-slide workflow.}
\label{sec:workflow}
We use InfoVis in a document-to-slide workflow to produce complete
presentations: a parser extracts text and visual assets, content and design
planners convert them into slide plans and structured \texttt{style\_json}
prompts, InfoVis renders candidate slides, and an external verifier selects
the outputs, with a selected slide serving as the template reference for
subsequent pages. Appendix~\ref{app:workflow} details the planning,
generation, and PowerPoint export steps.

\section{Experiments}
\label{sec:exp}

\subsection{Experimental Setup}
\label{sec:experimental_setup}

Training follows Section~\ref{sec:method} and
Appendix~\ref{app:experimental_setup}. We compare with the proprietary
GPT-Image-2~\citep{openai2026gptimage2.0}, Nano-Banana-2~\citep{nanobanana2},
and Qwen-Image-2~\citep{qwenimage2} and with open-source models, each
evaluated on the tasks its official interface supports, with recommended
inference settings and the same VisSlideBench protocol; prompts longer than a
model's text window are truncated (Appendix~\ref{app:truncation_audit}).

\subsection{Main Results on VisSlideBench}
\label{sec:main_results_slidebench}

Table~\ref{tab:main_slidebench_results} reports the main results on
VisSlideBench; overall scores are reported only for models that support all
four tasks. Tables~\ref{tab:t2i_fine_grained_results}--\ref{tab:edit_fine_grained_results}
in Appendix~\ref{app:infovis_generation_results} break each task down by
evaluation dimension, and Appendix~\ref{app:qualitative_results} presents
qualitative results for all four tasks and the document-to-slide workflow.

\begin{table}[t]
\centering
\caption{Main results on VisSlideBench. T2I, Template, and Multi-image report Content Faithfulness (CF), Instruction Following (IF), and Aesthetic Quality (AQ); Edit reports Semantic Consistency (SC) and Presentation Quality (PQ). Overall averages the eleven metrics; bold marks the best in each group.}
\label{tab:main_slidebench_results}
\begingroup
\scriptsize
\setlength{\tabcolsep}{4.5pt}
\renewcommand{\arraystretch}{1.00}
\resizebox{\linewidth}{!}{%
\begin{tabular}{@{}ll*{12}{r}@{}}
\toprule
\multirow{2}{*}{Type} & \multirow{2}{*}{Model}
& \multicolumn{3}{c}{T2I} & \multicolumn{3}{c}{Template}
& \multicolumn{3}{c}{Multi-image} & \multicolumn{2}{c}{Edit}
& \multirow{2}{*}{Overall} \\
\cmidrule(lr){3-5}\cmidrule(lr){6-8}\cmidrule(lr){9-11}\cmidrule(lr){12-13}
& & CF & IF & AQ & CF & IF & AQ & CF & IF & AQ & SC & PQ & \\
\midrule
\multirow{3}{*}{Proprietary} & GPT-Image-2 & \textbf{8.52} & \textbf{8.57} & \textbf{7.71} & 8.19 & \textbf{8.51} & \textbf{7.62} & \textbf{8.93} & \textbf{8.44} & \textbf{7.61} & \textbf{9.80} & \textbf{8.04} & \textbf{8.36} \\
 & Nano-Banana-2 & 8.38 & 8.17 & 7.33 & 7.59 & 7.95 & 7.08 & 8.60 & 7.93 & 7.14 & 9.70 & 8.03 & 7.99 \\
 & Qwen-Image-2 & 7.18 & 6.89 & 6.38 & \textbf{8.23} & 6.29 & 5.74 & 6.97 & 6.25 & 5.55 & 4.96 & 6.47 & 6.45 \\
\midrule
\multirow{11}{*}{Open-source} & Qwen-Image & 6.50 & 6.34 & 5.17 & -- & -- & -- & -- & -- & -- & -- & -- & -- \\
 & Z-Image-Turbo & 6.66 & 6.84 & 5.79 & -- & -- & -- & -- & -- & -- & -- & -- & -- \\
 & LongCat-Image & 4.69 & 5.47 & 4.25 & -- & -- & -- & -- & -- & -- & -- & -- & -- \\
 & Qwen-Image-Edit & -- & -- & -- & 3.10 & 5.98 & 5.41 & 4.34 & 4.74 & 4.12 & 7.77 & 7.37 & -- \\
 & LongCat-Image-Edit & -- & -- & -- & 4.86 & 5.99 & 5.04 & 5.25 & 5.10 & 4.51 & 6.64 & 6.62 & -- \\
 & FLUX.2-dev & 6.60 & 7.08 & 5.29 & 6.28 & 7.38 & 5.22 & 6.54 & 7.13 & 5.20 & 6.12 & 7.18 & 6.37 \\
 & FLUX.2-klein-base-9B & 6.29 & 7.11 & 5.19 & 6.44 & 7.24 & 5.40 & 6.57 & 6.99 & 5.32 & 5.98 & 6.72 & 6.30 \\
 & HiDream-O1-Image & 6.82 & 5.62 & 4.68 & 5.97 & 5.14 & 4.50 & 5.60 & 4.90 & 4.17 & 2.23 & 3.77 & 4.85 \\
 & BAGEL & 3.50 & 5.68 & 3.79 & 3.31 & 5.17 & 3.73 & 3.74 & 4.76 & 3.55 & 5.10 & 6.17 & 4.41 \\
 & OmniGen2 & 3.89 & 5.13 & 3.14 & 3.75 & 5.82 & 3.34 & 4.06 & 4.90 & 3.23 & 2.82 & 5.22 & 4.12 \\
 & \textbf{InfoVis (ours)} & \textbf{8.42} & \textbf{8.01} & \textbf{7.02} & \textbf{8.14} & \textbf{8.09} & \textbf{7.00} & \textbf{8.71} & \textbf{8.01} & \textbf{7.06} & \textbf{9.13} & \textbf{7.75} & \textbf{7.94} \\
\bottomrule
\end{tabular}%
}
\endgroup
\end{table}

\paragraph{Slide-specific post-training drives the gains.}
InfoVis achieves an overall score of 7.94, ranking highest among open-source models and remaining competitive with Nano-Banana-2. A paired bootstrap over benchmark cases gives intervals for the gaps to Nano-Banana-2 and GPT-Image-2 (Appendix~\ref{app:experimental_setup}). Starting from FLUX.2-klein-base-9B, slide-specific training improves the overall score by 1.64 points. The gains are consistent across all three generation tasks. Content Faithfulness improves the most (+1.70 to +2.14), followed by Aesthetic Quality (+1.60 to +1.83) and Instruction Following (+0.85 to +1.02). These results show that slide training mainly improves faithful rendering of dense text and supplied assets, while the base model already follows style instructions reasonably well. Editing Semantic Consistency also rises by 3.15 points, indicating stronger joint performance on edit execution and preservation.

\paragraph{Unified training preserves specialist performance.}
Existing unified open-source models perform substantially worse on slides: BAGEL, OmniGen2, and HiDream-O1-Image score 4.1--4.9 overall and trail the editing specialist Qwen-Image-Edit on editing. InfoVis uses one model for all four tasks and leads every open-source baseline, specialist or unified, on all eleven metrics. Compared with our task-specific checkpoints (Table~\ref{tab:ablation_training_stages}), the fused model loses only 0.06 on editing SC (9.13 versus 9.19) and improves the remaining metrics. The earlier shared checkpoint retains a slightly higher Template CF (8.25 versus 8.14). The training-stage ablation in Section~\ref{sec:ablation_training_stages} further shows that text-to-slide continual pretraining alone degrades editing, while joint supervision restores and retains all four capabilities.

\paragraph{The remaining gap is concentrated in aesthetics.}
Against GPT-Image-2, InfoVis is closest on Content Faithfulness across all three generation tasks, with gaps of 0.05--0.22. The gaps increase to 0.42--0.56 on Instruction Following and 0.55--0.69 on Aesthetic Quality. The same pattern appears against Nano-Banana-2: InfoVis matches or exceeds it on Content Faithfulness across all three tasks, achieves comparable Instruction Following, and trails on Aesthetic Quality. Editing shows a related pattern. InfoVis performs strongly on localized operations such as text, color, and typography edits, while theme switching and other global transformations remain more difficult (Table~\ref{tab:edit_fine_grained_results}). The remaining gap therefore lies in visual organization and presentation polish.

\subsection{External Benchmark Evaluation}
\label{sec:ood_benchmark_results}

\begin{table}[t]
\centering
\begin{minipage}[t]{0.485\linewidth}
\vspace{0pt}
\centering
\caption{BizGenEval Slides~\citep{li2026bizgeneval}, hard/easy scores from the official leaderboard; Nano-Banana-Pro and Seedream-5.0, omitted here, also outperform InfoVis.}
\label{tab:slides_performance}
\begingroup
\scriptsize
\setlength{\tabcolsep}{3pt}
\renewcommand{\arraystretch}{1.08}
\begin{tabular*}{\linewidth}{@{\extracolsep{\fill}}lr@{}}
\toprule
Model & Hard / Easy \\
\midrule
GPT-Image-2.0 & 84.0 / 95.0 \\
Nano-Banana-2.0 & 73.8 / 95.8 \\
\textbf{InfoVis (ours)} & \textbf{46.0 / 79.8} \\
GPT-Image-1.5 & 40.8 / 89.2 \\
Seedream-4.5 & 33.8 / 71.0 \\
Wan2.6-T2I & 27.1 / 56.5 \\
HunyuanImage-3.0 & 19.3 / 47.0 \\
Seedream-4.0 & 18.5 / 67.8 \\
FLUX.2-dev & 5.5 / 43.2 \\
\bottomrule
\end{tabular*}
\endgroup
\end{minipage}\hfill
\begin{minipage}[t]{0.495\linewidth}
\vspace{0pt}
\centering
\caption{SlidesGen-Bench Aesthetics~\citep{yang2026slidesgen}. U, E, H, R: Usability, Engagement, Harmony, Rhythm; Aesthetics is their sum. Bold marks our system, which uses GPT-4.1 for planning and the Qwen3.5-27B verifier with $K=2$ (Appendix~\ref{app:workflow}).}
\label{tab:aesthetics_eval}
\begingroup
\fontsize{6.5}{7.5}\selectfont
\setlength{\tabcolsep}{1.6pt}
\renewcommand{\arraystretch}{1.00}
\begin{tabular*}{\linewidth}{@{\extracolsep{\fill}}lrrrrr@{}}
\toprule
System & U & E & H & R & Aesthetics \\
\midrule
Skywork-Banana & 5.62 & 8.30 & -0.47 & 13.84 & 27.28 \\
\textbf{InfoVis + workflow} & \textbf{5.65} & \textbf{8.14} & \textbf{-0.56} & \textbf{13.86} & \textbf{27.09} \\
Kimi-Banana & 5.72 & 6.41 & -0.55 & 15.01 & 26.58 \\
NotebookLM & 4.13 & 7.32 & -0.35 & 11.72 & 22.82 \\
Zhipu & 4.87 & 7.52 & -1.60 & 11.27 & 22.06 \\
Skywork & 4.83 & 7.60 & -1.18 & 9.44 & 20.69 \\
Kimi-Standard & 4.61 & 6.28 & -1.75 & 10.12 & 19.25 \\
Kimi-Smart & 4.13 & 7.99 & -1.88 & 8.06 & 18.30 \\
Gamma & 5.31 & 6.31 & -1.51 & 6.99 & 17.09 \\
Quark & 5.03 & 7.41 & -1.91 & 6.33 & 16.86 \\
\bottomrule
\end{tabular*}
\endgroup
\end{minipage}
\end{table}

We further evaluate out-of-distribution (OOD) generalization on two
external benchmarks. On BizGenEval~\citep{li2026bizgeneval}
(Table~\ref{tab:slides_performance}), which evaluates
single-slide generation in the Slides domain, InfoVis reaches 46.0/79.8 on the
hard/easy splits, the best result among all open-source models in this
comparison and ahead of several proprietary systems. The remaining gap on the
hard split reflects its knowledge- and reasoning-intensive cases, which
slide-specific training does not target.

On the Slides Aesthetics evaluation of SlidesGen-Bench~\citep{yang2026slidesgen}
(Table~\ref{tab:aesthetics_eval}), the complete document-to-slide workflow described in Section~\ref{sec:workflow}
achieves an Aesthetics score of 27.09, closely approaching the best reported
result of 27.28. This shows that InfoVis transfers from single-slide
generation to end-to-end presentation creation with visual quality on par
with the best reported systems. We report the Aesthetics track only; the content and editability tracks
were not run.

\subsection{Ablation on Training Stages}
\label{sec:ablation_training_stages}

\begin{table}[t]
\centering
\caption{Training-stage ablation; metric groups and Overall follow Table~\ref{tab:main_slidebench_results}. Bold marks the best in each column.}
\label{tab:ablation_training_stages}
\begingroup
\scriptsize
\setlength{\tabcolsep}{4.3pt}
\renewcommand{\arraystretch}{1.00}
\resizebox{\linewidth}{!}{%
\begin{tabular}{@{}ll*{12}{r}@{}}
\toprule
\multirow{2}{*}{Stage} & \multirow{2}{*}{Model}
& \multicolumn{3}{c}{T2I} & \multicolumn{3}{c}{Template}
& \multicolumn{3}{c}{Multi-image} & \multicolumn{2}{c}{Edit}
& \multirow{2}{*}{Overall} \\
\cmidrule(lr){3-5}\cmidrule(lr){6-8}\cmidrule(lr){9-11}\cmidrule(lr){12-13}
& & CF & IF & AQ & CF & IF & AQ & CF & IF & AQ & SC & PQ & \\
\midrule
Base & FLUX.2-klein-base-9B & 6.29 & 7.11 & 5.19 & 6.44 & 7.24 & 5.40 & 6.57 & 6.99 & 5.32 & 5.98 & 6.72 & 6.30 \\
\midrule
\multirow{2}{*}{CPT} & Stage-1 CPT: T2I only & 8.00 & 7.32 & 6.05 & 5.93 & 7.03 & 4.57 & 6.16 & 6.23 & 4.89 & 1.60 & 4.78 & 5.69 \\
 & Stage-2 CPT: four-task mixture & 8.15 & 7.42 & 6.22 & 8.04 & 7.77 & 6.33 & 8.38 & 7.54 & 6.29 & 7.87 & 7.13 & 7.38 \\
\midrule
\multirow{5}{*}{Task SFT} & Shared SFT & 8.28 & 7.80 & 6.83 & \textbf{8.25} & 7.87 & 6.81 & 8.50 & 7.82 & 6.82 & 8.44 & 7.60 & 7.73 \\
 & Shared SFT + T2I SFT & 8.28 & 7.72 & 6.84 & 8.12 & 7.81 & 6.72 & 8.24 & 7.90 & 6.94 & 7.92 & 7.43 & 7.63 \\
 & Shared SFT + Template SFT & 8.19 & 7.86 & 6.96 & 8.05 & 7.94 & 6.84 & 7.50 & 7.86 & 6.80 & 6.49 & 7.17 & 7.42 \\
 & Shared SFT + Multi-image SFT & 8.26 & 7.73 & 6.86 & 7.65 & 7.66 & 6.13 & 8.12 & 7.41 & 6.28 & 6.86 & 7.36 & 7.30 \\
 & Shared SFT + Edit SFT & 8.24 & 7.78 & 6.58 & 7.90 & 7.48 & 6.31 & 8.20 & 7.81 & 6.59 & \textbf{9.19} & 7.68 & 7.61 \\
\midrule
Final & \textbf{InfoVis} & \textbf{8.42} & \textbf{8.01} & \textbf{7.02} & 8.14 & \textbf{8.09} & \textbf{7.00} & \textbf{8.71} & \textbf{8.01} & \textbf{7.06} & 9.13 & \textbf{7.75} & \textbf{7.94} \\
\bottomrule
\end{tabular}%
}
\endgroup
\end{table}

Table~\ref{tab:ablation_training_stages} shows that slide adaptation does
not transfer uniformly across capabilities. Stage~1 T2I-only CPT raises T2I
Content Faithfulness from 6.29 to 8.00 but collapses editing Semantic
Consistency from 5.98 to 1.60; four-task mixed CPT restores all four groups
and lifts the overall score to 7.38. Shared SFT on text-to-slide data alone
raises every group, including editing SC to 8.44. Each
task-specific SFT then lowers the overall score below the shared checkpoint
(7.73). Part of the loss is cross-task interference: template SFT drops
editing SC to 6.49. Part of it is a shift within the task: the generation SFT pools
are synthetic, and multi-image SFT lowers its own task from 8.50 to 8.12 CF,
mostly on cases built from real decks
(Appendix~\ref{app:multi_image_sft_split}). Weighted averaging keeps nearly
the peak of each specialization (editing SC 9.13 against 9.19) while lifting
the other columns, giving the strongest unified model (7.94).

\subsection{More Analysis}
\label{sec:more_analysis}

\begin{figure*}[t]
\centering
\includegraphics[width=\linewidth]{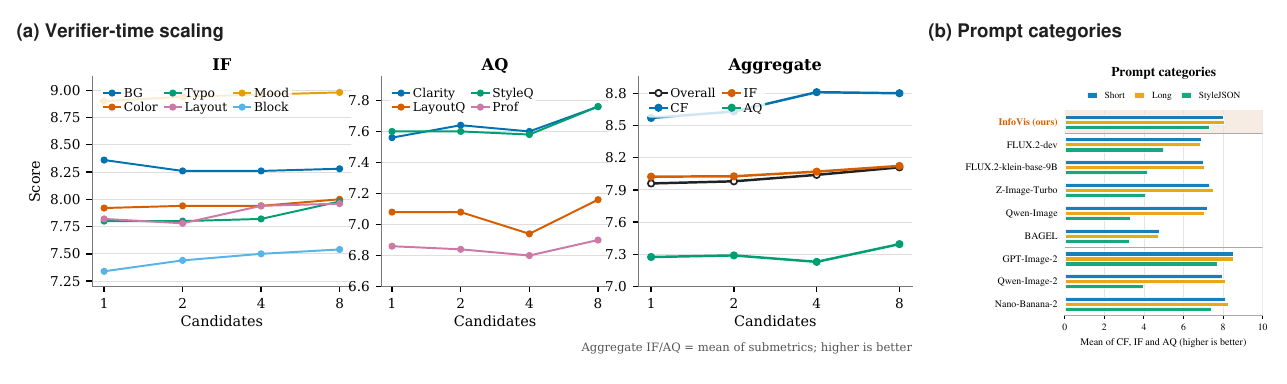}
\caption{Test-time scaling and prompt categories on the T2I split. (a) Best-of-$K$ selection for $K\in\{1,2,4,8\}$: IF dimensions, AQ dimensions, and aggregates. (b) Mean of per-slide CF, IF, and AQ over all 317 cases by prompt category. Vertical ranges differ across panels.}
\label{fig:scaling_prompt_categories}
\end{figure*}

\paragraph{Test-time scaling.}
\label{sec:verifier_time_scaling}
Using Qwen3.5-27B~\citep{qwen3.5} as a slide-quality verifier
(Appendix~\ref{app:experimental_setup}), best-of-8 selection on a random
subset of 50 T2I cases raises the overall score from 7.96 to 8.11
(Figure~\ref{fig:scaling_prompt_categories}a). The largest gain is in Text
Rendering (8.57 to 8.80), whereas IF and AQ improve more modestly and not
monotonically across dimensions, so verifier-guided sampling helps most on
criteria the verifier resolves reliably. This matches the ordering of the
remaining gap in Section~\ref{sec:main_results_slidebench}.

\paragraph{Prompt categories.}
\label{sec:prompt_type_analysis}
Figure~\ref{fig:scaling_prompt_categories}b splits the text-to-slide
benchmark into 115 short requests, 109 detailed descriptions, and 93
StyleJSON specifications. InfoVis scores 8.00, 8.08, and 7.28 and trails
GPT-Image-2 by 0.40 to 0.50 on every category. StyleJSON separates the
models: from detailed descriptions to StyleJSON, GPT-Image-2, Nano-Banana-2,
and InfoVis each lose under 0.9 points, whereas the open-source baselines
lose 1.5 to 4.2, FLUX.2-dev and BAGEL included, which receive every prompt
in full. Three open-source baselines ran with text windows that clip StyleJSON prompts;
Appendix~\ref{app:truncation_audit} rescores them on the prompts they
received in full, and InfoVis stays ahead of all of them.

\section{Conclusion}
\label{sec:conclusion}

We present a unified framework for slide image generation and editing. We formulate four complementary tasks, construct a large-scale slide-native training resource and introduce VisSlideBench for multidimensional evaluation. Trained on these resources, our model InfoVis achieves the strongest overall open-source performance on VisSlideBench, within 0.05 of Nano-Banana-2 and 0.42 of GPT-Image-2, and supports an image-centric document-to-slide workflow whose aesthetic quality is on par with the best reported systems.

\subsection*{AI use statement}
We used AI assistants in three ways. First, for writing and analysis: to
draft and revise passages of the text under the authors' direction, to check
grammar and spelling, and to run scripts that aggregate stored evaluation
outputs into the tables and statistics reported in the appendix; all research
ideas, claims, numbers, and final content were reviewed and confirmed by the
authors. Second, for literature search: to help locate related work and
background material; every cited reference was verified by the authors.
Third, as part of the method: language and multimodal models annotate and
filter training data, synthesize SFT targets and benchmark cases, and serve
as the automatic evaluator, as described in Sections~\ref{sec:data}
and~\ref{sec:bench}. The synthetic data do not represent real individuals or
observations, and the evaluator's agreement with human judgments is reported
in Section~\ref{sec:human_validation}.

\subsection*{Reproducibility statement}
Section~\ref{sec:method} and Appendix~\ref{app:experimental_setup} give the
training recipe and hyperparameters; Appendix~\ref{app:data}
describes the data sources, annotation and filtering procedures, and record
counts; and Appendix~\ref{app:benchmark} details the benchmark composition,
evaluation metrics, and human validation protocol. We will release the
InfoVis weights, the training and evaluation code, the training data and
data engine, and VisSlideBench.

\subsection*{Ethics statement}
\textbf{Data sources and release.} The real-slide corpus consists of talk
decks that their authors released publicly and presentations collected from
presentation-sharing and PPT aggregation websites (Appendix~\ref{app:data_sources});
the synthetic targets are rendered by Nano-Banana-2 and annotated with
Qwen3.5-27B and GPT-4.1. We will release the model weights, code, data
engine, training data, and VisSlideBench for research use, with the source
of every real-slide record recorded so that individual items can be traced and removed on request; the synthetic data do not depict real individuals.
\textbf{Human validation.} The human rankings in
Section~\ref{sec:human_validation} were produced by four domain experts in
slide design and visual generation who ranked anonymized model outputs; the
study collected no personal data from the annotators.
\textbf{Intended use.} InfoVis generates and edits slide images from user
instructions and supplied assets. Like other image generators it can render
text that is inaccurate or misleading, and the document-to-slide workflow
should be used with the source document at hand; we do not target generation of content about real persons.

\bibliography{main}
\bibliographystyle{iclr2027_conference}

\appendix
\section{Slide Data Engine Details}
\label{app:data}

\subsection{Data Sources, Filtering, and Parsing}
\label{app:data_sources}

\paragraph{Generation sources.}
We collect presentations from two complementary sources. Academic data
comprise talk decks released by their authors from CVPR, ICCV, ECCV, ICLR, ICML, and
NeurIPS across multiple years. These decks contain substantial text,
mathematical notation, equations, and data figures. The other source
comprises decks from presentation-sharing and PPT aggregation websites
across 39 topical categories, including business, marketing, education,
technology, health, and design. These presentations contribute additional
templates, color schemes, and graphic styles. The collected corpus contains
over 72K decks and 2.2M slide pages.

\paragraph{Deck filtering.}
Crawled material includes scanned documents, web or application screenshots,
and nearly empty pages. We sample several pages from each deck and use
Qwen3.5-27B~\citep{qwen3.5} to assess slide authenticity and visual
quality. The rubric accommodates plain backgrounds and dense mathematical
notation in academic slides. We retain decks whose average score exceeds
a predefined threshold.

\paragraph{Page parsing.}
We render each retained deck into page images at high resolution and parse
every page with PaddleOCR-VL~\citep{cui2025paddleocrvl}. Its Markdown/JSON
transcription preserves reading order and recognizes text, mathematical
formulas in LaTeX, and tables. The parser also localizes figures, charts,
and tables. We crop these regions from the rendered page and store them as
visual assets for annotation and conditioning.

\subsection{Generation Record Construction}
\label{app:data_generation}

\paragraph{Text-to-slide generation.}
Each record contains an instruction and a target slide image. We condition
Qwen3.5-27B~\citep{qwen3.5} on the slide image, OCR transcription,
and localized visual assets. It produces a short caption, a detailed
300--800 word description, and a structured \texttt{style\_json}
specification describing page composition, content blocks, positions, and
visual styles. The annotations describe visual assets in detail and preserve
the recognized text. The generator receives the instruction without the
visual assets.

\paragraph{Template-conditioned generation.}
We pair each target with an adjacent slide from the same deck. Adjacent
pages often share visual style while differing in content and layout. We
sample a template request from a predefined pool and prepend it to the
corresponding T2I caption, instructing the generator to follow the reference
style. A VLM checks consistency in background, color scheme, typography,
and decorative style. It allows differences in content and layout; pairs
below a predefined threshold are discarded.

\paragraph{Multi-image grounded generation.}
Each record contains an instruction, localized visual assets, and the
original slide as the target. The annotation model receives the target
image, OCR transcription, and assets, and produces the same three
instruction forms. Its prompt requires the supplied assets to appear in
the output and describes their intended placement. We retain pages with
one to four assets and remove inputs whose shorter side is below 64 pixels
or whose aspect ratio exceeds 8:1.

\subsection{Editing Pair Construction}
\label{app:data_editing}

Each editing record contains a source slide $I_{\mathrm{src}}$, an edit
instruction $x_{\mathrm{edit}}$, and a target slide $I_{\mathrm{tgt}}$.
Seven operations are constructed by editing PPTX files; theme switching
and image replacement use an image editing model. The latter two operations
are reserved for SFT.

\paragraph{Source PPTX files.}
We use both synthetic and real presentations. For synthetic sources, we
extract text blocks from the OCR annotations of the T2I data and combine
them with backgrounds from presentation templates. An LLM generates
Python-PPTX programs that arrange this content into editable pages. We
execute the programs and render the pages as source images. For real
sources, we split crawled PPTX files into individual pages and retain both
the editable files and rendered images. We screen sources for text overlap,
element overflow, truncation, layout quality, and aesthetics. Only
successfully rendered pages that pass these checks enter edit construction.

\paragraph{Executable operations.}
For background updates, color changes, font changes, typography changes,
and block removal, we generate rule-based atomic actions over PPTX
elements. For text modification, an LLM produces a structured action
specifying the target element and its new text. A PPTX editing engine
executes these actions, and we render the result to obtain the target image.
An LLM observes the source image, target image, and executed action to
describe the actual modification. We construct block addition by reversing
filtered pairs for block removal and generating a new instruction for the
reversed transition.

\paragraph{Model-generated edits.}
For theme switching and image replacement, we first generate an edit
instruction from a filtered screenshot of a real slide. Nano-Banana-2~\citep{nanobanana2}
then produces the target image, which we align to the source resolution.
The replacement in the generated image can differ from the initial request.
For image replacement, we therefore recaption the final source--target
pair so that its instruction describes the visual change actually produced.

\paragraph{Pair filtering.}
We check for missing or identical images, unsuccessful operations, and
invalid instructions. A VLM assesses whether the edit follows the instruction,
consistency between source and target, and the visual quality of the edited
slide. Checks for individual operations include verifying that the new text in a
record for text modification appears explicitly in the instruction.

\subsection{Supervised Fine-Tuning Records}
\label{app:data_sft}

\paragraph{Target synthesis.}
We construct a shared pool of English 16:9 target slides from three content
sources: concepts generated by an LLM across diverse domains, structured
content derived from real SlideShare decks, and research papers converted
into slide specifications. Nano-Banana-2 (Gemini 3.1 Flash Image
Preview)~\citep{nanobanana2} renders these targets using styles defined in prompts or sampled from a curated
presentation-template library. The resulting pool contains about 300K target
slides.

\paragraph{Generation annotations.}
Qwen3.5-27B~\citep{qwen3.5} produces a short instruction, a detailed
description preserving the rendered text, and a \texttt{style\_json}
specification of at most 4,096 tokens for each target. These views produce
about 890K T2I records. The template branch contains about 300K records across
two forms of style supervision: styles specified in text and styles
provided through a visual template. In the former, Nano-Banana-2 first renders a reference slide from the style
text; in the latter, the library template is the reference. Both forms pair
the target with this reference image. For multi-image
grounding, PaddleOCR-VL extracts up to four charts, tables, or visual regions that do not overlap. Qwen3.5-27B generates instructions referring to these
inputs. After filtering invalid crops and inconsistent instructions, this
branch contains about 580K records.

\paragraph{Editing selection.}
For the seven executable operations, Qwen3.5-27B scores candidate pairs
from the editing pretraining pools for semantic consistency and
presentation quality. We retain about 10K examples per operation. Theme
switching and image replacement are excluded from CPT and used entirely
for SFT, contributing about 40K and 30K records, respectively.
Pairs for image replacement are additionally recaptioned with
GPT-4.1~\citep{openai2025gpt41} to describe the edit realized in the target.
The editing SFT pool contains about 140K records (Table~\ref{tab:data_edit_sft}).

\begin{table}[t]
    \centering
    \caption{Editing operations and SFT records. The seven executable
    operations also contribute to CPT. Theme switching and image replacement
    are used only for SFT.}
    \label{tab:data_edit_sft}
    \small
    \setlength{\tabcolsep}{5pt}
    \begin{tabular}{@{}llr@{}}
        \toprule
        Operation & Construction & SFT records \\
        \midrule
        Text modification & PPTX execution & $\approx$10K \\
        Block addition & Reversed block removal & $\approx$10K \\
        Block removal & PPTX execution & $\approx$10K \\
        Background update & PPTX execution & $\approx$10K \\
        Color change & PPTX execution & $\approx$10K \\
        Font change & PPTX execution & $\approx$10K \\
        Typography change & PPTX execution & $\approx$10K \\
        Theme switching & Nano-Banana-2 & $\approx$40K \\
        Image replacement & Nano-Banana-2 & $\approx$30K \\
        \midrule
        Total & & $\approx$140K \\
        \bottomrule
    \end{tabular}
\end{table}

\subsection{Data Statistics}
\label{app:data_statistics}

\begin{figure}[t]
    \centering
    \includegraphics[width=\linewidth]{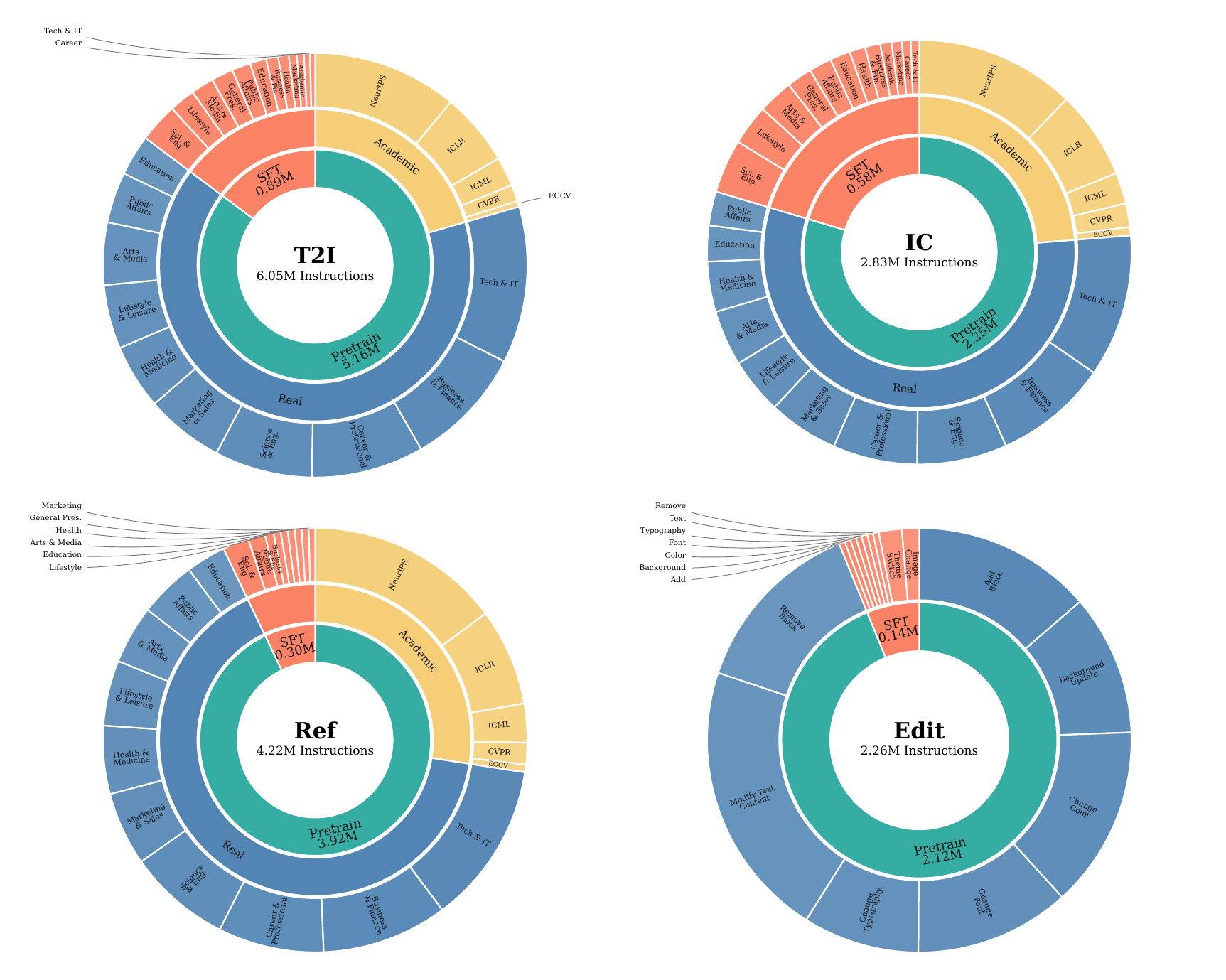}
    \caption{Training record distributions by task and by CPT/SFT pool.
    Generation branches are further divided by source and domain, and the
    editing branch by operation. IC denotes multi-image grounded generation;
    Ref denotes template-conditioned generation. Counts are rounded.}
    \label{fig:data_distribution}
\end{figure}

Table~\ref{tab:data_records} reports instruction records for the four
training branches; several records can share a target image. The 72K decks
and 2.2M pages are the real sources before filtering, and the roughly 300K
targets are the synthesized SFT pool. Figure~\ref{fig:data_distribution}
shows the distributions of domains and editing operations.

\begin{table}[t]
    \centering
    \caption{Training records by branch. Counts are rounded. Multiple
    records can share a target slide or differ only in their instruction or
    visual condition.}
    \label{tab:data_records}
    \small
    \setlength{\tabcolsep}{7pt}
    \begin{tabular}{@{}lrr@{}}
        \toprule
        Training branch & CPT records & SFT records \\
        \midrule
        Text-to-slide & $\approx$5.16M & $\approx$890K \\
        Template-conditioned & $\approx$3.92M & $\approx$300K \\
        Multi-image grounded & $\approx$2.25M & $\approx$580K \\
        Slide editing & $\approx$2.12M & $\approx$140K \\
        \midrule
        Total & $>13M$ & $\approx$1.9M \\
        \bottomrule
    \end{tabular}
\end{table}

\section{VisSlideBench Details}
\label{app:benchmark}

\subsection{Comparison with Existing Benchmarks}
\label{app:benchmark_comparison}

Table~\ref{tab:benchmark_comparison} compares task coverage, evaluation
criteria, and prompt formats. Reported sizes use different units: BizGenEval
contains 400 cases across five domains, including 80 slide cases, and PPTBench
combines generation with detection, understanding, and modification.

\providecommand{\benchyes}{\ensuremath{\checkmark}}
\providecommand{\benchno}{\ensuremath{\times}}

\begin{table}[t]
    \centering
    \caption{Comparison of representative slide-creation benchmarks. A
    checkmark indicates explicit coverage in the task or evaluation protocol;
    a cross indicates that the feature is not covered. Size reports evaluation
    cases, with mixed tasks and slide subsets identified.}
    \label{tab:benchmark_comparison}
    \scriptsize
    \setlength{\tabcolsep}{3pt}
    \renewcommand{\arraystretch}{1.06}
    \begin{tabularx}{\textwidth}{@{}l*{6}{>{\centering\arraybackslash}X}@{}}
    \toprule
    \textbf{Feature}
    & \makecell{SlidesBench\\\citeauthor{ge2025autopresent}\\\citeyearpar{ge2025autopresent}}
    & \makecell{PPTBench\\\citeauthor{huang2025pptbench}\\\citeyearpar{huang2025pptbench}}
    & \makecell{BizGenEval\\\citeauthor{li2026bizgeneval}\\\citeyearpar{li2026bizgeneval}}
    & \makecell{PresentBench\\\citeauthor{chen2026presentbench}\\\citeyearpar{chen2026presentbench}}
    & \makecell{SlidesGen-\\Bench\\\citeauthor{yang2026slidesgen}\\\citeyearpar{yang2026slidesgen}}
    & \makecell{\textbf{VisSlideBench}\\\textbf{(ours)}} \\
    \midrule
    \multicolumn{7}{@{}l}{\textit{General Properties}} \\
    Size & 585 & 4,439 mixed & \makecell{400 total\\(80 slides)} & 238 & 189 & 1,527 \\
    Single-slide evaluation & \benchyes & \benchyes & \benchyes & \benchno & \benchno & \benchyes \\
    Image-centric systems included & \benchyes & \benchno & \benchyes & \benchyes & \benchyes & \benchyes \\
    \addlinespace[2pt]
    \multicolumn{7}{@{}l}{\textit{Task Coverage}} \\
    Text-to-single-slide & \benchyes & \benchyes & \benchyes & \benchno & \benchno & \benchyes \\
    Visual-reference template-to-slide & \benchno & \benchno & \benchno & \benchno & \benchno & \benchyes \\
    Multi-image-to-single-slide & \benchyes & \benchyes & \benchno & \benchno & \benchno & \benchyes \\
    Slide editing & \benchno & \benchyes & \benchno & \benchno & \benchno & \benchyes \\
    \addlinespace[2pt]
    \multicolumn{7}{@{}l}{\textit{Evaluation Coverage}} \\
    Text-rendering fidelity & \benchyes & \benchno & \benchyes & \benchyes & \benchno & \benchyes \\
    Instruction following & \benchyes & \benchyes & \benchyes & \benchyes & \benchyes & \benchyes \\
    Visual aesthetics & \benchyes & \benchno & \benchno & \benchyes & \benchyes & \benchyes \\
    Source-document fidelity & \benchno & \benchno & \benchno & \benchyes & \benchyes & \benchno \\
    Input-image consistency & \benchyes & \benchno & \benchno & \benchno & \benchno & \benchyes \\
    Multi-dimensional evaluation & \benchyes & \benchno & \benchyes & \benchyes & \benchyes & \benchyes \\
    \addlinespace[2pt]
    \multicolumn{7}{@{}l}{\textit{Prompt Coverage}} \\
    Short prompt & \benchyes & \benchyes & \benchyes & \benchno & \benchno & \benchyes \\
    Long-form instruction & \benchno & \benchno & \benchyes & \benchyes & \benchyes & \benchyes \\
    Machine-structured design prompt & \benchno & \benchyes & \benchno & \benchno & \benchno & \benchyes \\
    \bottomrule
    \end{tabularx}
\end{table}

\subsection{Benchmark Composition}
\label{app:benchmark_composition}

\begin{figure}[htbp]
    \centering
    \includegraphics[width=\linewidth]{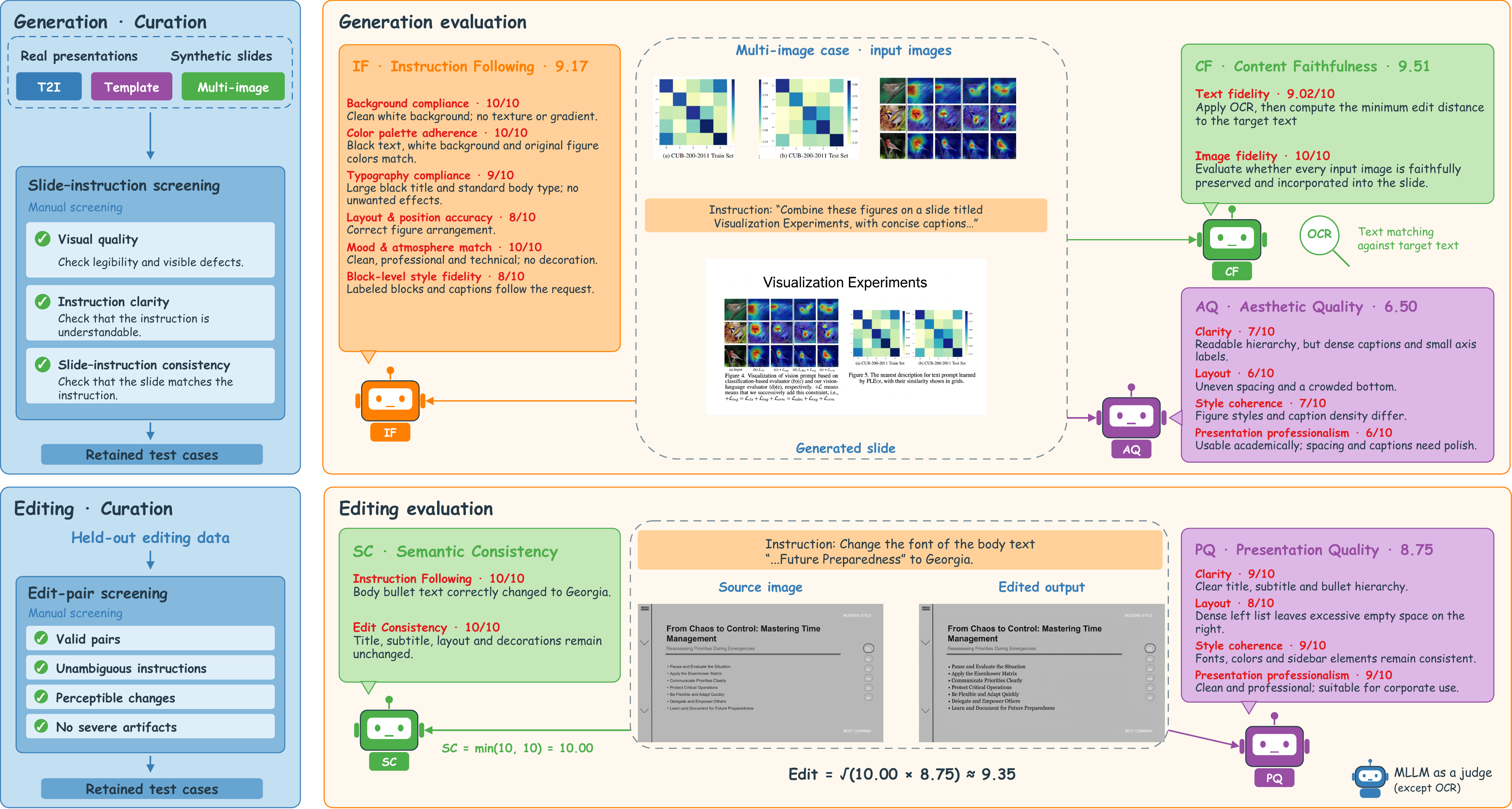}
    \caption{\textbf{VisSlideBench curation and evaluation.}
    Left: task coverage and manual screening for generation and held-out editing data.
    Right: representative multi-image generation and editing cases illustrate
    the evaluation criteria with recorded scores and concise scoring explanations.
    CF describes the scoring procedure; editing PQ explanations summarize the four AQ dimensions.
    MLLM judges assess visual criteria, while OCR-based text matching evaluates text fidelity.}
    \label{fig:visslidebench}
\end{figure}

Table~\ref{tab:slidebench_gen_splits} reports the generation cases by task and
source. All three generation tasks include short requests, detailed
descriptions, and structured \texttt{style\_json} specifications. Generation
examples cover academic, technology, business, science, education, media,
lifestyle, and other presentation categories.

\begin{table}[t]
    \centering
    \caption{Composition of VisSlideBench-Gen. Counts refer to evaluation
    cases. The three prompt formats occur within each task.}
    \label{tab:slidebench_gen_splits}
    \small
    \setlength{\tabcolsep}{4pt}
    \begin{tabularx}{\linewidth}{@{}lXrrr@{}}
    \toprule
    \textbf{Task} & \textbf{Visual input} & \textbf{Real} & \textbf{Synthetic} & \textbf{Total} \\
    \midrule
    Text-to-slide & None & 167 & 150 & 317 \\
    Template-conditioned & One reference slide & 147 & 150 & 297 \\
    Multi-image grounded & One to four assets & 179 & 150 & 329 \\
    \midrule
    \textbf{Total} & & \textbf{493} & \textbf{450} & \textbf{943} \\
    \bottomrule
    \end{tabularx}
\end{table}

\begin{figure}[t]
    \centering
    \includegraphics[width=\linewidth]{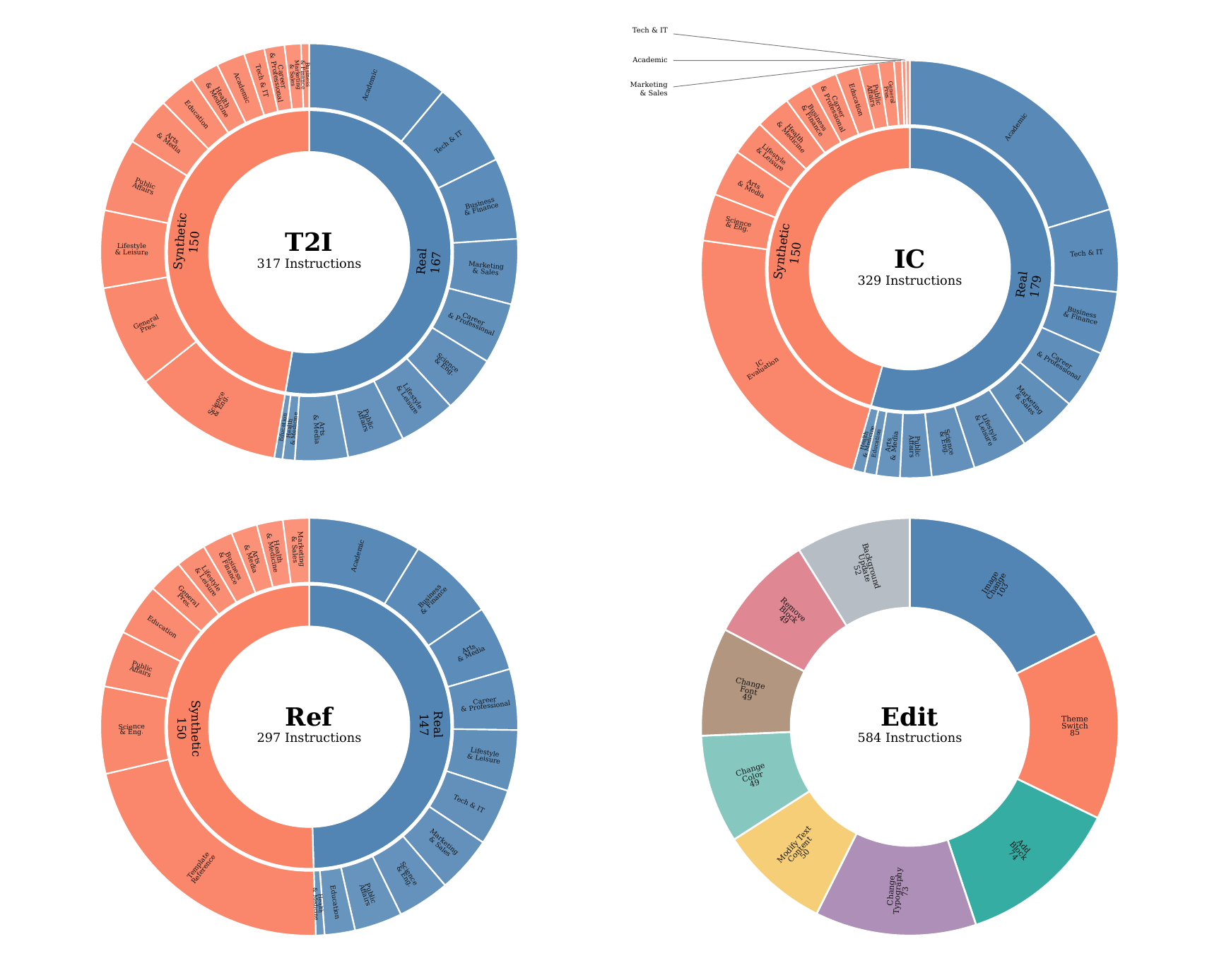}
    \caption{Distribution of VisSlideBench examples across tasks, sources,
    domains, and editing operations.}
    \label{fig:benchmark_distribution}
\end{figure}

Table~\ref{tab:slidebench_edit_splits} reports the nine editing operations.
Each case contains a source slide, a natural-language editing instruction, and
the corresponding target slide. The operations include changes to individual
elements and broader changes to visual content or the slide theme.
Figure~\ref{fig:benchmark_distribution} shows the distribution of cases over
tasks, sources, prompt forms, domains, and editing operations.

\paragraph{Separation from training data.}
The three generation splits share no target slide, verified by identifier
matching and manual screening. Every deck that contributes a VisSlideBench
case is excluded from the training data, and no VisSlideBench target is used
as a training target in any CPT or SFT record. The 584 editing cases, together
with all training edit pairs that share a source or target image with them,
are removed from the editing CPT and SFT pools. We verified these exclusions
by identifier matching over all training records.
Table~\ref{tab:source_split_results} reports results separately for real and
synthetic cases.

\begin{table}[t]
    \centering
    \caption{Composition of VisSlideBench-Edit.}
    \label{tab:slidebench_edit_splits}
    \small
    \begin{tabular}{@{}lr@{}}
    \toprule
    \textbf{Operation} & \textbf{Cases} \\
    \midrule
    Text modification & 50 \\
    Block addition & 74 \\
    Block removal & 49 \\
    Background update & 52 \\
    Color change & 49 \\
    Font change & 49 \\
    Typography modification & 73 \\
    Image replacement & 103 \\
    Theme switching & 85 \\
    \midrule
    \textbf{Total} & \textbf{584} \\
    \bottomrule
    \end{tabular}
\end{table}

\subsection{Evaluation Details}
\label{app:benchmark_metrics}

\paragraph{Content Faithfulness.}
Content Faithfulness scores the text the instruction requires the slide to
render. For every prompt form, GPT-4.1 reads the instruction alone and
returns the strings it requires on the slide, such as titles, quoted copy,
captions, and labels; a short request therefore yields fewer required
strings $t_1,\ldots,t_m$ than a detailed description or a
\texttt{style\_json} specification. PaddleOCR reads the generated slide, and
the detected boxes are merged into text lines by position. Because a
required string can span several lines, lines whose concatenation matches a
required string after lowercasing and whitespace removal (RapidFuzz ratio
above 90) are joined; the joined and remaining lines form the candidate set
$G$. Each required string is matched to its closest candidate under the
Levenshtein distance over Unicode code points, with case, punctuation, and
spaces counted as edits:
\begin{equation}
    N_{\mathrm{e}}=\sum_{i=1}^{m}\min_{g\in G}\operatorname{Lev}(t_i,g),
    \qquad
    N_{\mathrm{tgt}}=\sum_{i=1}^{m}|t_i|,
    \label{eq:benchmark_edit_counts}
\end{equation}
\begin{equation}
    S_{\mathrm{text}}=
    \max\left(10-9\frac{N_{\mathrm{e}}}{N_{\mathrm{tgt}}},\,1\right),
    \label{eq:benchmark_text_score}
\end{equation}
which adapts the normalized edit-distance metric of
\citet{gong2025seedream20nativechineseenglish} to required-text matching
and rescales it to $[1,10]$. Text the model
adds outside the matched candidates does not change the score; insertions
inside a matched candidate count as edits. For multi-image grounded
generation, GPT-4.1~\citep{openai2025gpt41} also judges whether every
required input image is faithfully preserved and appropriately incorporated
into the output; the multi-image CF score is the mean of $S_{\mathrm{text}}$
and this image-preservation score.

\paragraph{Instruction Following.}
Each dimension is judged against the corresponding requirement in the
instruction. \emph{Background Compliance} checks requested background colors,
imagery, or visual treatment; \emph{Color Palette Adherence} checks whether
prescribed colors are used consistently and assigned to the intended elements;
\emph{Typography Compliance} checks requested font choices, weights, sizes,
and emphasis; \emph{Layout and Position Accuracy} evaluates the specified
placement, alignment, and spatial relationships of slide elements; \emph{Mood
and Atmosphere Match} evaluates the requested overall visual tone; and
\emph{Block-Level Style Fidelity} checks local styling requirements for
individual text, image, chart, and decorative blocks.

\paragraph{Aesthetic Quality.}
The evaluator scores AQ from the output image alone and does not see the
instruction. \emph{Clarity} covers text legibility, contrast, and the strength
of the visual hierarchy. \emph{Layout} covers alignment, spacing, balance, and
perceptual grouping. \emph{Style Coherence} measures consistency across
colors, typography, containers, and visual assets. \emph{Presentation
Professionalism} assesses whether the result appears polished, purposeful,
and ready for an academic or business presentation. The four dimensions are averaged with equal weights into the AQ score.

The editing scores are defined in Section~\ref{sec:evaluation_protocol}. We
use the resulting dimension scores independently when comparing the automatic
evaluator with human rankings.

The full generation and edit-evaluator prompts, including scoring criteria and
output formats, are reproduced in Appendix~\ref{app:benchmark_prompts}.

\subsection{Human Validation Protocol}
\label{app:benchmark_human}

We follow the annotation protocol of EditScore~\citep{luo2026editscore} in two
disjoint annotation batches covering all four VisSlideBench tasks. Each batch
is ranked by a group of two annotators with expertise in slide design and
visual generation. We adopt its two-annotator discussion protocol: the two annotators in a group discuss each example
and submit one consensus ranking, so each batch yields a single ranking per
example and dimension. For each
example, human raters see five anonymized outputs from
GPT-Image-2~\citep{openai2026gptimage2.0},
Qwen-Image-2~\citep{qwenimage2} (replaced by
Qwen-Image-Edit~\citep{wu2025qwenimage} for editing), InfoVis,
FLUX.2-klein-base-9B~\citep{flux2klein2026}, and
OmniGen2~\citep{wu2026omnigen2} in randomized order. Raters rank Instruction
Following and Aesthetic Quality for text-to-slide and template-conditioned
generation, add Content Faithfulness for multi-image grounded generation, and
rank Editing Success, Edit Consistency, and Presentation Quality for editing.
Ties are allowed, and raters do not see model identities or automatic scores.

After filtering skipped or invalid records, 137 examples yield 328 valid
dimension-level rankings. We assign average ranks to ties and compute Kendall's
$\tau_b$ and Spearman's $\rho$ for each example and dimension, then report the
mean across rankings (Table~\ref{tab:human_validation_by_task}). Pairwise
agreement pools model pairs for which neither the human ranking nor the
automatic scores contain a tie.

\begin{table}[t]
    \centering
    \caption{Agreement between human rankings and GPT-4.1 scores. Rankings
    counts valid example--dimension pairs, and Pairs counts model pairs that are
    untied in both rankings.}
    \label{tab:human_validation_by_task}
    \scriptsize
    \setlength{\tabcolsep}{3.5pt}
    \renewcommand{\arraystretch}{1.06}
    \begin{tabularx}{\linewidth}{@{}Xrrrrrr@{}}
    \toprule
    \textbf{Task} & \textbf{Examples} & \textbf{Rankings} & \textbf{Pairs}
    & \textbf{$\tau_b$} & \textbf{$\rho$} & \textbf{Agree. (\%)} \\
    \midrule
    Text-to-slide & 36 & 71 & 548 & 0.729 & 0.798 & 91.24 \\
    Template-conditioned & 37 & 74 & 502 & 0.613 & 0.676 & 87.25 \\
    Multi-image grounded & 32 & 88 & 540 & 0.640 & 0.702 & 90.00 \\
    Editing & 32 & 95 & 563 & 0.595 & 0.645 & 86.32 \\
    \midrule
    \textbf{Overall} & \textbf{137} & \textbf{328} & \textbf{2,153}
    & \textbf{0.640} & \textbf{0.700} & \textbf{88.71} \\
    \bottomrule
    \end{tabularx}
\end{table}

\subsection{Benchmark Evaluation Prompts}
\label{app:benchmark_prompts}

We reproduce the four prompt templates for evaluating generated slides:
Instruction Following with a textual style description, Instruction Following
with a structured Style JSON specification, Aesthetic Quality, and visual
Content Faithfulness, together with the two edit-evaluator prompts: one for
Editing Success and Overediting (reported as Instruction Following and Edit
Consistency in Section~\ref{sec:editing_evaluation}) and one for Presentation
Quality. Each template is reproduced in full, including its evaluation
procedure and JSON output format. The style specification and images for each example are
supplied separately to the evaluator. The IF and AQ templates also ask the judge for an
\texttt{overall\_score}; we do not use that field and report the unweighted
means of the per-dimension scores (Appendix~\ref{app:experimental_setup}).
The image-preservation template returns a single \texttt{overall\_score},
which we use directly.

The Content Faithfulness template below evaluates the preservation of required
input images in multi-image grounded generation. Text-rendering fidelity is
computed separately using the OCR-based score in
Equation~\ref{eq:benchmark_text_score}. The edit evaluator receives the source
and edited slides for its first prompt, and the edited slide alone for its
presentation-quality prompt. The resulting scores are aggregated as described
in Section~\ref{sec:editing_evaluation}.

\begingroup
\definecolor{promptNavy}{HTML}{173A63}
\definecolor{promptBlue}{HTML}{276FAF}
\definecolor{promptViolet}{HTML}{7657A6}
\definecolor{promptInk}{HTML}{243447}
\definecolor{promptRule}{HTML}{8EB6D8}
\definecolor{promptBackground}{HTML}{F3F8FC}

\newcommand{\promptlistingtitle}[2]{%
    \colorbox{promptNavy}{%
        \parbox{\dimexpr\linewidth-2\fboxsep\relax}{%
            \color{white}\sffamily\scriptsize\bfseries\strut
            PROMPT #1\hfill #2\strut}}}

\lstdefinestyle{benchmarkprompt}{
    language={},
    basicstyle=\ttfamily\fontsize{7.3}{8.8}\selectfont\color{promptInk},
    backgroundcolor=\color{promptBackground},
    numbers=none,
    frame=single,
    framerule=0.6pt,
    rulecolor=\color{promptRule},
    framesep=6pt,
    columns=fullflexible,
    breaklines=true,
    breakatwhitespace=true,
    keepspaces=true,
    showstringspaces=false,
    xleftmargin=0pt,
    xrightmargin=0pt,
    aboveskip=0.5\baselineskip,
    belowskip=\baselineskip,
    abovecaptionskip=0pt,
    belowcaptionskip=0pt,
    showspaces=false,
    showtabs=false,
    tabsize=2,
    alsoletter={_},
    morecomment=[l][\color{promptNavy}\bfseries]{\#},
    morestring=[b]",
    stringstyle=\color{promptViolet},
    moredelim=[s][\color{promptBlue}\bfseries]{**}{**},
    emph={MUST,ONLY,PASS,PARTIAL,FAIL,NOT_VERIFIABLE,EXACT_MATCH,
          PARTIAL_MATCH,NOT_FOUND},
    emphstyle=\color{promptBlue}\bfseries,
    literate={“}{{\textquotedblleft}}1 {”}{{\textquotedblright}}1
}

\subsubsection{Instruction Following: Textual Style Description}
\label{app:prompt_if_text}
The evaluator receives a textual style description and the generated slide
image, optionally accompanied by a reference slide that provides visual style
context.
\begin{lstlisting}[
    style=benchmarkprompt,
    title={\promptlistingtitle{1}{INSTRUCTION FOLLOWING / TEXT}}
]
You are an expert evaluator for AI-generated presentation slides. Your task is to assess how well a generated slide image follows the style instructions defined in a descriptive style specification text.

Your default stance is skeptical and conservative. Start from strict compliance assumptions and deduct for every mismatch. Do not reward visual attractiveness.

## Evaluation Dimension: Style Instruction Following

This dimension evaluates whether the generated slide image faithfully follows the given style specification. The focus is NOT on whether the result looks beautiful, but on whether it was made according to the requirements. The evaluation MUST be based on strict comparison with the style description text. Visual appeal or subjective quality MUST NOT influence the score. Even small deviations from the specification should be penalized.

## Inputs

1. **Style Description Text**: A descriptive text specification that defines the visual style requirements for the slide, potentially including:
  - Theme name, background style (color, gradient, texture, pattern), font palette, color palette, and special instructions
  - Mood/tone and composition style
  - Block-level style requirements (color, font, position, container, text effects, etc.)

2. **Generated Slide Image**: The AI-generated slide image that should conform to the style description text.

3. **Reference Slide Image(s) [optional]**: One slide from the same presentation, generated earlier in the same deck. Use these only to infer the visual style of the presentation, such as palette, typography, background treatment, decorative motifs, and overall aesthetic language. Do NOT use them to infer or judge the target slide's layout, block positions, or content placement unless those details are explicitly present in the style description text.

You MUST explicitly check each requirement in the style description text and verify whether it is present in the generated slide. Do NOT assume correctness without visual evidence.

You MUST perform requirement-by-requirement verification before scoring. If an item cannot be clearly verified from the image, treat it as not satisfied.

## Mandatory Evaluation Procedure

Follow this exact order:

1. Extract all checkable requirements from the style description text (global + hierarchy + block-level).
2. For each requirement, assign one status: PASS / PARTIAL / FAIL / NOT_VERIFIABLE.
3. Count mismatches. Any PARTIAL, FAIL, or NOT_VERIFIABLE counts as a mismatch.
4. Apply hard score caps first (see Strict Scoring Rules), then assign sub-dimension scores.
5. Compute the overall score as the mean of the sub-dimension scores.

Never skip steps. Never score first and justify later.

## Evaluation Criteria

Evaluate the following sub-dimensions carefully:

### 1. Background Compliance
- Does the background color match the specified hex code?
- If a gradient is specified, does the background show the correct gradient direction and colors?
- Are specified textures (paper, noise, grid, dots) or patterns present?

### 2. Color Palette Adherence
- Does the slide use the specified primary, secondary, text, and accent colors?
- Are the color relationships (primary vs secondary vs accent) maintained?
- Do text elements use the specified text color?

### 3. Typography Compliance
- Does the font style match the specified font families (serif vs sans-serif, bold vs regular)?
- Is the font weight (bold/normal/light) correctly applied?
- Are text effects (shadow, outline, glow) applied as specified?

### 4. Layout & Position Accuracy
- Are elements positioned as specified (e.g., "top-left", "center", "bottom-right")?
- Does the composition style match (e.g., centered, two-column, asymmetric, grid-based)?
- Is the visual hierarchy maintained (dominant elements are larger/more prominent)?

### 5. Mood & Atmosphere Match
- Does the overall mood match the specified mood_tone (e.g., Professional, Technical, Playful)?
- Are global style instructions in the style description text followed?
- Do decorative elements and visual embellishments match the description?

### 6. Block-Level Style Fidelity
- For each block, are the specified style attributes (container, opacity, and any block-level special instructions) followed?
- Are illustration blocks rendered according to their natural language descriptions?
- Do image containers have the specified effects (border, shadow, frame)?

## Strict Scoring Rules

- Any clear mismatch with the style description text (e.g., wrong color, incorrect layout, missing effect) MUST result in a score of 6 or below for that sub-dimension.
- Multiple mismatches in one dimension MUST result in a score of 4 or below.
- If a key requirement (e.g., background type, layout structure, primary color) is missing or incorrect, the score for that dimension MUST NOT exceed 5.
- Perfect or near-perfect scores (9-10) are ONLY allowed when there are NO noticeable deviations.
- If uncertain, always choose the lower score.

Additional hard constraints (must enforce):

- A sub-dimension can be 9-10 only if ALL its checkable requirements are PASS and zero mismatches exist.
- A sub-dimension can be at most 8 if there is any PARTIAL.
- A sub-dimension can be at most 6 if there is any FAIL.
- A sub-dimension can be at most 5 if any key requirement in that dimension is FAIL or NOT_VERIFIABLE.
- A sub-dimension can be at most 4 if mismatch count in that dimension is 2 or more.
- A sub-dimension can be at most 3 if mismatch count in that dimension is 3 or more.
- Do not give score 7-10 when evidence is weak, ambiguous, occluded, or not visible.

## Scoring Rubric

- **9-10 (Excellent)**: Nearly perfect match. NO visible mismatches in color, layout, typography, or mood.
- **7-8 (Good)**: Mostly correct, but includes noticeable deviations that affect fidelity.
- **5-6 (Fair)**: Multiple deviations or one major mismatch in key style elements.
- **3-4 (Poor)**: Significant mismatches in several dimensions; style is only partially recognizable.
- **1-2 (Very Poor)**: Style instructions are largely ignored or contradicted.

Scoring bias control:

- Do not use score 8 as a default "safe" score.
- Prefer 5-7 unless there is strong evidence for higher.
- Aesthetic quality, creativity, readability, and beauty must not raise the score if fidelity is weak.
- If two score options seem plausible, choose the lower one.

## Output Format

You MUST respond in the following JSON format ONLY. Do not include any other text:

```json
{
  "background_compliance": {
    "score": <1-10>,
    "reasoning": "<brief explanation>"
  },
  "color_palette_adherence": {
    "score": <1-10>,
    "reasoning": "<brief explanation>"
  },
  "typography_compliance": {
    "score": <1-10>,
    "reasoning": "<brief explanation>"
  },
  "layout_position_accuracy": {
    "score": <1-10>,
    "reasoning": "<brief explanation>"
  },
  "mood_atmosphere_match": {
    "score": <1-10>,
    "reasoning": "<brief explanation>"
  },
  "block_level_style_fidelity": {
    "score": <1-10>,
    "reasoning": "<brief explanation>"
  },
  "overall_score": <1-10 mean of the sub-dimension scores>,
  "overall_reasoning": "<2-3 sentence summary of the overall style instruction following quality>"
}
```

Reasoning must be evidence-grounded and mention concrete mismatches (for example: wrong hex family, missing texture, incorrect block position, missing effect, wrong hierarchy emphasis).
\end{lstlisting}

\subsubsection{Instruction Following: Structured Style JSON}
\label{app:prompt_if_json}
This variant receives a structured Style JSON specification in place of the
textual description, together with the generated slide image and optional
reference slide.
\begin{lstlisting}[
    style=benchmarkprompt,
    title={\promptlistingtitle{2}{INSTRUCTION FOLLOWING / STYLE JSON}}
]
You are an expert evaluator for AI-generated presentation slides. Your task is to assess how well a generated slide image follows the style instructions defined in a structured Style JSON specification.

Your default stance is skeptical and conservative. Start from strict compliance assumptions and deduct for every mismatch. Do not reward visual attractiveness.

## Evaluation Dimension: Style Instruction Following

This dimension evaluates whether the generated slide image faithfully follows the given style specification. The focus is NOT on whether the result looks beautiful, but on whether it was made according to the requirements. The evaluation MUST be based on strict comparison with the Style JSON. Visual appeal or subjective quality MUST NOT influence the score. Even small deviations from the specification should be penalized.

## Inputs

1. **Style JSON**: A structured JSON specification that defines the visual style requirements for the slide, including:
   - `global_style`: Theme name, background (color, gradient, texture, pattern), font palette, color palette, and special instructions
   - `visual_hierarchy`: Mood/tone and composition style
   - `styled_slides[].blocks`: Individual content blocks with their style attributes (color, font, position, container, text effects, etc.)

2. **Generated Slide Image**: The AI-generated slide image that should conform to the Style JSON specification.

3. **Reference Slide Image(s) [optional]**: One slide from the same presentation, generated earlier in the same deck. Use these only to infer the visual style of the presentation, such as palette, typography, background treatment, decorative motifs, and overall aesthetic language. Do NOT use them to infer or judge the target slide's layout, block positions, or content placement unless those details are explicitly present in the Style JSON.

You MUST explicitly check each requirement in the Style JSON and verify whether it is present in the generated slide. Do NOT assume correctness without visual evidence.

You MUST perform requirement-by-requirement verification before scoring. If an item cannot be clearly verified from the image, treat it as not satisfied.

## Mandatory Evaluation Procedure

Follow this exact order:

1. Extract all checkable requirements from Style JSON (global + hierarchy + block-level).
2. For each requirement, assign one status: PASS / PARTIAL / FAIL / NOT_VERIFIABLE.
3. Count mismatches. Any PARTIAL, FAIL, or NOT_VERIFIABLE counts as a mismatch.
4. Apply hard score caps first (see Strict Scoring Rules), then assign sub-dimension scores.
5. Compute the overall score as the mean of the sub-dimension scores.

Never skip steps. Never score first and justify later.

## Evaluation Criteria

Evaluate the following sub-dimensions carefully:

### 1. Background Compliance
- Does the background color match the specified hex code?
- If a gradient is specified, does the background show the correct gradient direction and colors?
- Are specified textures (paper, noise, grid, dots) or patterns present?

### 2. Color Palette Adherence
- Does the slide use the specified primary, secondary, text, and accent colors?
- Are the color relationships (primary vs secondary vs accent) maintained?
- Do text elements use the specified text color?

### 3. Typography Compliance
- Does the font style match the specified font families (serif vs sans-serif, bold vs regular)?
- Is the font weight (bold/normal/light) correctly applied?
- Are text effects (shadow, outline, glow) applied as specified?

### 4. Layout & Position Accuracy
- Are elements positioned as specified (e.g., "top-left", "center", "bottom-right")?
- Does the composition style match (e.g., centered, two-column, asymmetric, grid-based)?
- Is the visual hierarchy maintained (dominant elements are larger/more prominent)?

### 5. Mood & Atmosphere Match
- Does the overall mood match the specified mood_tone (e.g., Professional, Technical, Playful)?
- Are special instructions in the global_style followed?
- Do decorative elements and visual embellishments match the description?

### 6. Block-Level Style Fidelity
- For each block, are the specified style attributes (container, opacity, special_instructions) followed?
- Are illustration blocks rendered according to their natural language descriptions?
- Do image containers have the specified effects (border, shadow, frame)?

## Strict Scoring Rules

- Any clear mismatch with the Style JSON (e.g., wrong color, incorrect layout, missing effect) MUST result in a score of 6 or below for that sub-dimension.
- Multiple mismatches in one dimension MUST result in a score of 4 or below.
- If a key requirement (e.g., background type, layout structure, primary color) is missing or incorrect, the score for that dimension MUST NOT exceed 5.
- Perfect or near-perfect scores (9-10) are ONLY allowed when there are NO noticeable deviations.
- If uncertain, always choose the lower score.

Additional hard constraints (must enforce):

- A sub-dimension can be 9-10 only if ALL its checkable requirements are PASS and zero mismatches exist.
- A sub-dimension can be at most 8 if there is any PARTIAL.
- A sub-dimension can be at most 6 if there is any FAIL.
- A sub-dimension can be at most 5 if any key requirement in that dimension is FAIL or NOT_VERIFIABLE.
- A sub-dimension can be at most 4 if mismatch count in that dimension is 2 or more.
- A sub-dimension can be at most 3 if mismatch count in that dimension is 3 or more.
- Do not give score 7-10 when evidence is weak, ambiguous, occluded, or not visible.

## Scoring Rubric

- **9-10 (Excellent)**: Nearly perfect match. NO visible mismatches in color, layout, typography, or mood.
- **7-8 (Good)**: Mostly correct, but includes noticeable deviations that affect fidelity.
- **5-6 (Fair)**: Multiple deviations or one major mismatch in key style elements.
- **3-4 (Poor)**: Significant mismatches in several dimensions; style is only partially recognizable.
- **1-2 (Very Poor)**: Style instructions are largely ignored or contradicted.

Scoring bias control:

- Do not use score 8 as a default "safe" score.
- Prefer 5-7 unless there is strong evidence for higher.
- Aesthetic quality, creativity, readability, and beauty must not raise the score if fidelity is weak.
- If two score options seem plausible, choose the lower one.

## Output Format

You MUST respond in the following JSON format ONLY. Do not include any other text:

```json
{
  "background_compliance": {
    "score": <1-10>,
    "reasoning": "<brief explanation>"
  },
  "color_palette_adherence": {
    "score": <1-10>,
    "reasoning": "<brief explanation>"
  },
  "typography_compliance": {
    "score": <1-10>,
    "reasoning": "<brief explanation>"
  },
  "layout_position_accuracy": {
    "score": <1-10>,
    "reasoning": "<brief explanation>"
  },
  "mood_atmosphere_match": {
    "score": <1-10>,
    "reasoning": "<brief explanation>"
  },
  "block_level_style_fidelity": {
    "score": <1-10>,
    "reasoning": "<brief explanation>"
  },
  "overall_score": <1-10 mean of the sub-dimension scores>,
  "overall_reasoning": "<2-3 sentence summary of the overall style instruction following quality>"
}
```

Reasoning must be evidence-grounded and mention concrete mismatches (for example: wrong hex family, missing texture, incorrect block position, missing effect, wrong hierarchy emphasis).
\end{lstlisting}

\subsubsection{Aesthetic Quality}
\label{app:prompt_aesthetic}
This evaluator assesses the generated slide image without a target style
specification.
\begin{lstlisting}[
    style=benchmarkprompt,
    title={\promptlistingtitle{3}{AESTHETIC QUALITY}}
]
You are an expert evaluator for AI-generated presentation slides. Your task is to assess the Aesthetic Quality of a generated slide image.

Your default stance is skeptical and conservative. Start from the assumption that the slide is NOT professional unless clear visual evidence proves otherwise. Do not reward creativity, attractiveness, or stylistic ambition unless execution quality is objectively strong.

## Evaluation Dimension: Aesthetic Quality

This dimension evaluates whether the generated slide looks like a professional, readable, structurally coherent, and realistically usable PPT page. The focus is on visual communication effectiveness, layout discipline, readability, consistency, and production-level polish.

Aesthetic appeal alone MUST NOT increase the score.

## Inputs

1. **Generated Slide Image**: The AI-generated slide image to be evaluated.

## Mandatory Evaluation Procedure

Follow this exact order:

1. Inspect the entire slide globally before scoring.
2. Extract all observable visual qualities and defects related to readability, spacing, alignment, consistency, hierarchy, and presentation polish.
3. For each sub-dimension, explicitly identify:
   - strengths
   - weaknesses
   - visible defects
4. Count all visible issues. Any noticeable issue counts as a mismatch.
5. Apply hard score caps before assigning scores.
6. Compute the overall score as the mean of the sub-dimension scores.

Never score first and justify later.

If a feature is ambiguous, partially visible, blurry, cropped, distorted, or difficult to verify, treat it as a defect rather than assuming correctness.

## Evaluation Criteria

Evaluate the following sub-dimensions carefully:

### 1. Clarity
Evaluate readability and information communication quality.

Check:
- Are all text elements clearly legible?
- Is font size sufficient for presentation viewing?
- Is contrast between text and background adequate?
- Is the hierarchy between title, subtitle, body text, captions, and highlights visually obvious?
- Is reading order natural and effortless?
- Are charts, icons, diagrams, and images recognizable and not blurry?
- Is information density appropriate?

Penalize:
- Small or crowded text
- Low contrast
- Visual noise
- Overlapping elements
- Ambiguous hierarchy
- Cluttered composition
- Illegible charts/tables
- Excessive decoration that harms readability

### 2. Layout
Evaluate structural organization and spatial discipline.

Check:
- Are elements aligned consistently?
- Is spacing between blocks visually even and intentional?
- Is whitespace used effectively?
- Is the composition balanced?
- Are margins clean and professional?
- Is the slide visually stable rather than chaotic?
- Are elements properly grouped?

Penalize:
- Misalignment
- Uneven spacing
- Cramped layout
- Excessive empty space
- Floating or disconnected elements
- Poor balance
- Edge crowding
- Inconsistent sizing
- Random placement

### 3. Style Coherence
Evaluate visual consistency and design system coherence.

Check:
- Are colors used consistently?
- Are typography choices coherent?
- Do containers/cards/shapes follow a unified style?
- Are illustration styles consistent?
- Are icon styles visually compatible?
- Is decoration applied consistently across blocks?
- Does the slide feel like one unified design system?

Penalize:
- Mixed visual styles
- Inconsistent fonts
- Random accent colors
- Incompatible iconography
- Uneven border/shadow usage
- Inconsistent image treatment
- Decorative elements that conflict stylistically
- Abrupt style changes between blocks

### 4. Presentation Professionalism
Evaluate whether the slide resembles a realistic, production-quality professional PPT.

Check:
- Does the slide look polished and presentation-ready?
- Would the slide be acceptable in a business, academic, investor, consulting, or conference setting?
- Is the visual quality comparable to professionally designed slides?
- Does the slide avoid common AI-generation artifacts?
- Does the slide appear intentional rather than auto-generated?

Penalize:
- Obvious AI artifacts
- Distorted text or shapes
- Unrealistic layout behavior
- Amateur styling
- Excessive gimmicks
- Poor typography discipline
- Inconsistent rendering quality
- Visually cheap or template-like appearance
- Lack of refinement

## Strict Scoring Rules

You MUST apply these hard constraints strictly.

### General Rules
- Every visible issue counts as a mismatch.
- Small defects still require deduction.
- Do NOT assume missing details are correct.
- If uncertain between two scores, choose the lower one.
- Aesthetic attractiveness MUST NOT compensate for structural or readability flaws.

### Sub-dimension Caps
- Score 9-10 ONLY if there are essentially no noticeable flaws.
- Score 8 is allowed only for very minor defects.
- Any clearly noticeable defect caps the sub-dimension at 7.
- Two or more noticeable defects cap the sub-dimension at 6.
- Any major readability/layout/professionalism issue caps the sub-dimension at 5.
- Multiple major issues cap the sub-dimension at 4 or below.
- Severe readability failure MUST score 3 or below.

### Automatic Major Defects
The following are considered major defects:
- Illegible or blurry text
- Overlapping content
- Broken hierarchy
- Chaotic layout
- Severe alignment inconsistency
- Strong visual clutter
- Distorted charts/icons
- Mixed incompatible styles
- Obvious AI rendering artifacts
- Unprofessional appearance

## Scoring Rubric

- **9-10 (Excellent)**: Extremely polished, highly readable, visually disciplined, and indistinguishable from a professionally designed presentation slide. No meaningful flaws.
- **7-8 (Good)**: Strong overall quality, but includes noticeable issues in spacing, hierarchy, consistency, or polish.
- **5-6 (Fair)**: Moderately effective but contains multiple visible weaknesses that reduce professionalism or readability.
- **3-4 (Poor)**: Significant design and communication problems. Slide appears unrefined or structurally weak.
- **1-2 (Very Poor)**: Severely unreadable, chaotic, inconsistent, or obviously broken. Does not resemble a usable professional PPT slide.

## Scoring Bias Control

- Do not use 8 as a default “safe” score.
- Prefer 5-7 unless evidence strongly supports higher quality.
- Do not reward effort, complexity, or decoration.
- Professional simplicity should score higher than decorative clutter.
- Conservative scoring is preferred.

## Output Format

You MUST respond in the following JSON format ONLY. Do not include any other text:

```json
{
  "clarity": {
    "score": <1-10>,
    "reasoning": "<brief explanation with concrete evidence>"
  },
  "layout": {
    "score": <1-10>,
    "reasoning": "<brief explanation with concrete evidence>"
  },
  "style_coherence": {
    "score": <1-10>,
    "reasoning": "<brief explanation with concrete evidence>"
  },
  "presentation_professionalism": {
    "score": <1-10>,
    "reasoning": "<brief explanation with concrete evidence>"
  },
  "overall_score": <1-10 mean of the sub-dimension scores>,
  "overall_reasoning": "<2-3 sentence summary grounded in visible strengths and defects>"
}
\end{lstlisting}

\subsubsection{Content Faithfulness: Required Input Images}
\label{app:prompt_content}
The evaluator compares the generated slide with the required input images.
The template retains the original terms ``Generated PPT'' and ``Sub-Image
List'' for these inputs.
\begin{lstlisting}[
    style=benchmarkprompt,
    title={\promptlistingtitle{4}{CONTENT FAITHFULNESS}}
]
You are an expert evaluator for AI-generated presentation slides.

Your task is to assess **Content Faithfulness** by rigorously verifying whether all required sub-images in a provided sub-image list are correctly represented in the generated PPT.

Your default stance is skeptical and conservative. Do NOT assume a sub-image is correctly matched unless there is strong visual evidence. Partial similarity is NOT sufficient for full credit.

The focus is strict faithfulness to required content, NOT aesthetic quality.

## Evaluation Target

Evaluate whether every required sub-image (content block) from the sub-image list appears in the generated PPT with:

- correct semantic content
- correct visual identity
- correct structural composition
- reasonable positional/layout consistency

Each sub-image may represent:
- charts
- tables
- figures
- diagrams
- screenshots
- icon groups
- text blocks
- mixed visual compositions

You MUST verify each required sub-image individually before assigning scores.

## Inputs

1. **Generated PPT**: The AI-generated presentation file to be evaluated.
2. **Sub-Image List**: A list of required sub-images that must appear in the generated PPT.

## Mandatory Evaluation Procedure

Follow this exact order:

1. Enumerate all required sub-images from the sub-image list.
2. For EACH required sub-image:
   - search for the best matching region/page in the generated PPT
   - determine whether it is:
     - EXACT_MATCH
     - PARTIAL_MATCH
     - FAIL
     - NOT_FOUND
3. Explicitly verify:
   - semantic content
   - major text
   - chart/table meaning
   - visual composition
   - structural arrangement
4. Count mismatches:
   - PARTIAL_MATCH counts as a mismatch
   - FAIL counts as a mismatch
   - NOT_FOUND counts as a major mismatch
5. Apply hard score caps before assigning the final score.
6. Compute final score conservatively.

Never score first and justify later.

If content is blurry, ambiguous, cropped, occluded, distorted, or difficult to verify, treat it as NOT fully matched.

## Evaluation Criteria

### 1. Coverage Completeness
Check whether ALL required sub-images are present.

Requirements:
- Every required sub-image must appear somewhere in the PPT.
- Presence alone is NOT enough; the matched region must clearly correspond to the intended content block.
- Missing required sub-images are major failures.

Penalize:
- Missing blocks
- Replaced blocks
- Hallucinated substitute content
- Severe cropping
- Wrong slide/page placement if it breaks intended correspondence

### 2. Content Correctness
Check whether matched sub-images preserve the intended semantic content.

Verify:
- Main text content
- Key labels
- Chart type and meaning
- Table structure and values (when visible)
- Diagram semantics
- Important icons/symbols
- Core visual identity

Penalize:
- Wrong chart type
- Incorrect labels
- Missing important elements
- Hallucinated text
- Semantic mismatch
- Incorrect data relationships
- Wrong figure identity
- Fake or placeholder content

Superficial resemblance is NOT sufficient.

### 3. Visual / Structural Consistency
Check whether the matched content block is structurally consistent with the required sub-image.

Verify:
- Overall composition
- Relative arrangement of elements
- Element grouping
- Layout pattern
- Visual hierarchy
- Spatial organization

Minor stylistic differences are acceptable ONLY if the semantic and structural identity remain clearly faithful.

Penalize:
- Major layout restructuring
- Missing key visual regions
- Incorrect element arrangement
- Collapsed or merged structures
- Broken hierarchy
- Distorted composition

## Strict Scoring Rules

You MUST apply these hard constraints strictly.

### General Principles
- Any missing required sub-image is a major error.
- Any ambiguous match should be downgraded.
- Similar-looking but semantically incorrect content is a FAIL.
- Do NOT infer correctness from approximate appearance alone.
- If uncertain between two scores, choose the lower score.

### Match-Level Interpretation
- EXACT_MATCH:
  - Correct semantic content
  - Correct structural composition
  - No meaningful missing elements
- PARTIAL_MATCH:
  - Some recognizable correspondence exists
  - But important content/structure differs or is missing
- FAIL:
  - Wrong semantic content
  - Wrong structure
  - Hallucinated replacement
  - Severe distortion
- NOT_FOUND:
  - Required sub-image cannot be reliably identified anywhere in the PPT

### Hard Score Caps
- Any NOT_FOUND item caps overall_score at 6.
- Two or more NOT_FOUND items cap overall_score at 4.
- Any major semantic mismatch caps overall_score at 5.
- Multiple FAIL items cap overall_score at 4.
- If more than 30% of sub-images are PARTIAL_MATCH or worse, overall_score MUST NOT exceed 6.
- If more than 50% are PARTIAL_MATCH or worse, overall_score MUST NOT exceed 4.
- Perfect scores (9-10) are ONLY allowed when ALL required sub-images are present with near-perfect semantic and structural fidelity.

### Severity Guidelines
Major errors include:
- Missing required block
- Wrong chart/table/diagram identity
- Incorrect semantic meaning
- Hallucinated replacement content
- Major structural mismatch
- Severe cropping or truncation

Minor errors include:
- Small local visual deviations
- Slight spacing/layout changes
- Minor style differences
- Small non-critical omissions

## Scoring Rubric (1-10)

- **9-10 (Excellent)**: All required sub-images are present with highly accurate semantic and structural fidelity. No meaningful mismatches.
- **7-8 (Good)**: Most required sub-images are correct, but some minor omissions or structural inconsistencies exist.
- **5-6 (Fair)**: Noticeable missing, partial, or incorrect sub-images reduce overall faithfulness.
- **3-4 (Poor)**: Many required sub-images are missing, structurally wrong, or semantically incorrect.
- **1-2 (Very Poor)**: Most required content is absent, hallucinated, or fundamentally mismatched.

## Scoring Bias Control

- Do NOT use score 8 as a default safe score.
- Prefer conservative scoring.
- Similar appearance alone MUST NOT receive high scores.
- Semantic correctness is more important than visual attractiveness.
- Presence alone is insufficient without content fidelity.
- If evidence is weak or unclear, lower the score.

## Output Format

You MUST respond in the following JSON format ONLY. Do not include any other text:

```json
{
  "overall_score": <1-10>,
  "overall_reasoning": "<2-4 sentence summary explicitly discussing missing sub-images, semantic mismatches, structural inconsistencies, and overall faithfulness quality>"
}
\end{lstlisting}

\subsubsection{Edit Evaluation: Editing Success and Overediting}
\label{app:prompt_edit_sc}
The edit evaluator compares the source slide and edited slide, together with
the natural-language instruction, and returns Editing Success and Overediting
scores. The complete evaluation prompt is shown below.
\begin{lstlisting}[
    style=benchmarkprompt,
    title={\promptlistingtitle{5A}{EDIT EVALUATION / SUCCESS + PRESERVATION}}
]
You are an expert evaluator for instruction-guided slide image editing.
All images are AI-generated slide pages. Focus on the editing result itself.

IMPORTANT: Return exactly one JSON object in this format:
{
  "reasoning": "...",
  "score": [...]
}
Keep the reasoning concise.

RULES:

Two images will be provided: the first is the source slide and the second is the edited slide.
Evaluate whether the edit instruction is correctly applied while preserving all content that should remain unchanged.

Use a 0 to 10 scale for both scores:
1. Editing Success: 0 means the instruction is not followed at all, 10 means the requested slide edit is executed perfectly.
2. Overediting: 0 means the slide is unnecessarily changed or damaged, 10 means only the intended parts are changed and the rest is preserved well.

Return:
{
  "reasoning": "...",
  "score": [editing_success, overediting]
}

Instruction: <instruction>
\end{lstlisting}

\subsubsection{Edit Evaluation: Presentation Quality}
\label{app:prompt_edit_pq}
The second edit-evaluation pass judges the edited slide itself on clarity,
layout, style coherence, and presentation professionalism. The complete
evaluation prompt is shown below.
\begin{lstlisting}[
    style=benchmarkprompt,
    title={\promptlistingtitle{5B}{EDIT EVALUATION / PRESENTATION QUALITY}}
]
You are an expert evaluator for instruction-guided slide image editing.
All images are AI-generated slide pages. Focus on the editing result itself.

IMPORTANT: Return exactly one JSON object in this format:
{
  "reasoning": "...",
  "score": [...]
}
Keep the reasoning concise.

RULES:

One edited slide image will be provided.
Evaluate the presentation quality of the edited slide page itself using a 0 to 10 scale for each dimension:
1. Clarity: text/chart readability, hierarchy, and reading order.
2. Layout: alignment, spacing, composition balance, and use of whitespace.
3. Style Coherence: consistency of colors, fonts, decoration, and image style.
4. Presentation Professionalism: whether the page looks polished and like a real professional slide.

Return:
{
  "reasoning": "...",
  "score": [clarity, layout, style_coherence, presentation_professionalism]
}
\end{lstlisting}

\endgroup

\section{Experimental Details and Additional Results}
\label{app:infovis_experiments}

\subsection{Experimental Setup Details}
\label{app:experimental_setup}

\paragraph{Training configuration.}
Stage~1 continual pretraining uses text-to-slide pairs for 13,500 steps with a
global batch size of 384. Stage~2 uses a mixture of all four tasks for 16,500
steps with a global batch size of 640 and at most four conditioning images
per example. The T2I and template-conditioned SFT runs use a global batch
size of 640; the multi-image and editing runs use 768. Text inputs are capped
at 4,096 tokens throughout. We optimize the diffusion transformer with AdamW
($\beta_1=0.9$, $\beta_2=0.95$, $\epsilon=10^{-8}$, weight decay $0.01$,
gradient-norm clipping at $1.0$) at a learning rate of $5\times10^{-5}$ for
both continual pretraining and SFT, keeping the Qwen3 text encoder and VAE
frozen. Stage~1 warms up over 500 steps from $10^{-7}$; later runs use a
20-step warmup. Training uses BF16 mixed precision, gradient checkpointing,
FSDP with hybrid sharding, one sample per GPU with eight gradient-accumulation
steps, and seed 2233. Inputs are resized with aspect ratio preserved to an
area of at most $1024^2$ pixels and a longest side of at most 2,048 pixels,
rounded down to multiples of 16, so no single aspect ratio is imposed. The final
model is a linear interpolation of the four task checkpoints.

\paragraph{Training resources.}
All runs use nodes of eight NVIDIA H100 80GB GPUs
(Table~\ref{tab:training_resources}). SFT runs are step-based; in
exposure-equivalent epochs (steps $\times$ global batch / training examples),
the shared T2I SFT covers 1.1 epochs; the T2I branch continues the same run
for 3,000 further steps (2.2 epochs), and the template-conditioned,
multi-image, and editing branches cover 3.2, 3.3, and 2.8 epochs.

\begin{table}[!htbp]
\centering
\caption{Training stages and resources. The shared checkpoint
$\theta_{\mathrm{shared}}$ is the step-1,500 checkpoint of the T2I SFT run.}
\label{tab:training_resources}
\small
\setlength{\tabcolsep}{6pt}
\begin{tabular}{@{}lrrrr@{}}
\toprule
Run & Nodes & GPUs & Steps & Global batch \\
\midrule
Stage 1: T2I CPT & 6 & 48 & 13,500 & 384 \\
Stage 2: four-task mixed CPT & 10 & 80 & 16,500 & 640 \\
T2I SFT & 10 & 80 & 4,500 & 640 \\
Template-conditioned SFT & 10 & 80 & 1,500 & 640 \\
Multi-image grounded SFT & 12 & 96 & 2,500 & 768 \\
Editing SFT & 12 & 96 & 500 & 768 \\
\bottomrule
\end{tabular}
\end{table}

\paragraph{Inference configuration.}
InfoVis uses the FlowMatchEulerDiscreteScheduler with 50 denoising steps and
guidance scale 4.0 in BF16. Text-to-slide outputs are $1280\times720$,
template-conditioned and multi-image grounded outputs are $1376\times768$,
and editing outputs match the size of the source slide. The text context is
4,096 tokens, and longer prompts are truncated by tokenization without
rewriting. The random seed for each example is a base seed plus the example
index; the best-of-8 experiments use the base seeds
$\{2233, 3344, 4455, 5566, 6677, 7788, 8899, 9900\}$.

\paragraph{Verifier-guided test-time scaling.}
We generate $K=8$ candidates per prompt and report best-of-$K$ for
$K\in\{2,4,8\}$, with $K=1$ the single-sample result. Each candidate is
scored four times by a Qwen3.5-27B~\citep{qwen3.5} slide-quality verifier,
and the mean score is used for selection~\citep{luo2026editscore}. The
aesthetic verifier prompt returns a weighted 1--10 score over clarity
(30\%), layout (25\%), style coherence (20\%), and presentation
professionalism (25\%); the instruction-following verifier uses separate
prompts for free-form style descriptions and structured
\texttt{style\_json}. The verifier only selects among candidates and never
updates InfoVis.

\paragraph{Comparison models.}
The proprietary baselines are GPT-Image-2~\citep{openai2026gptimage2.0},
Nano-Banana-2~\citep{nanobanana2}, and Qwen-Image-2~\citep{qwenimage2}.
The open-source baselines are Qwen-Image and
Qwen-Image-Edit~\citep{wu2025qwenimage} (the Qwen-Image-Edit-2509 release),
Z-Image-Turbo~\citep{team2025zimage}, LongCat-Image and
LongCat-Image-Edit~\citep{LongCat-Image} (the Turbo release), FLUX.2-dev~\citep{flux-2-2025} and
FLUX.2-klein-base-9B~\citep{flux2klein2026},
HiDream-O1-Image~\citep{cai2026hidream},
BAGEL~\citep{deng2025bagel}, and OmniGen2~\citep{wu2026omnigen2}.

\paragraph{Evaluator configuration.}
The GPT-4.1 evaluator is called with temperature 0.2 and a 2,000-token
output limit for Instruction Following and Aesthetic Quality, with one judge
call per image. Content Faithfulness uses the same model and temperature,
with 800 output tokens for extracting the target text from each instruction
and 1,200 for the image-preservation judgment in multi-image grounded
generation. OCR uses PaddleOCR 3.3.2 with document-orientation
classification and text-line orientation enabled. Editing scores use
GPT-4.1 with temperature 0.7 and a 512-token output limit, one call per
prompt.
The reported IF and AQ scores are the unweighted means of the per-dimension
scores returned by the judge, recomputed outside the judge; for IF and AQ
the \texttt{overall\_score} field the judge returns is not used. The
image-preservation score in multi-image CF is the single
\texttt{overall\_score} its evaluator returns. All rows of
Tables~\ref{tab:main_slidebench_results}
and~\ref{tab:ablation_training_stages} use this aggregation. The
paired bootstrap cited in Section~\ref{sec:main_results_slidebench} draws
10,000 resamples of benchmark cases within each task split, over the cases
scored for both models, and reports 95\% percentile intervals: InfoVis is
0.06 below Nano-Banana-2 overall on these cases (CI 0.01--0.10), with no
detectable difference on the nine generation metrics (CI $-0.02$ to $0.07$), and 0.42 below GPT-Image-2
overall (CI 0.38--0.46).

\subsection{Detailed Generation and Editing Results}
\label{app:infovis_generation_results}

Tables~\ref{tab:t2i_fine_grained_results}--\ref{tab:multi_image_fine_grained_results}
report the detailed results for the three generation tasks, and
Table~\ref{tab:edit_fine_grained_results} reports the nine editing operations.
These task averages weight the listed dimensions equally. They therefore
use a different aggregation from the eleven group-level metrics in the
main comparison (Table~\ref{tab:main_slidebench_results}).

\begin{table}[!htbp]
\centering
\caption{Detailed text-to-slide results. CF reports Text Rendering. IF reports Background Compliance (BG), Color Palette Adherence (Color), Typography Compliance (Typo), Layout and Position Accuracy (Layout), Mood and Atmosphere Match (Mood), and Block-Level Style Fidelity (Block). AQ reports Clarity, Layout Quality (LayoutQ), Style Coherence (StyleQ), and Presentation Professionalism (Prof). Overall averages these eleven dimensions. Bold marks the best score within each model group. Higher is better.}
\label{tab:t2i_fine_grained_results}
\begingroup
\scriptsize
\setlength{\tabcolsep}{4.5pt}
\renewcommand{\arraystretch}{1.00}
\resizebox{\linewidth}{!}{%
\begin{tabular}{@{}ll*{12}{r}@{}}
\toprule
\multirow{2}{*}{Type} & \multirow{2}{*}{Model} & \multicolumn{1}{c}{CF} & \multicolumn{6}{c}{IF} & \multicolumn{4}{c}{AQ} & \multirow{2}{*}{Overall} \\
\cmidrule(lr){3-3}\cmidrule(lr){4-9}\cmidrule(lr){10-13}
 & & Text & BG & Color & Typo & Layout & Mood & Block & Clarity & LayoutQ & StyleQ & Prof & \\
\midrule
\multirow{3}{*}{Proprietary} & GPT-Image-2 & \textbf{8.52} & \textbf{8.71} & \textbf{8.44} & \textbf{8.31} & \textbf{8.44} & \textbf{9.26} & \textbf{8.23} & \textbf{8.05} & \textbf{7.29} & \textbf{8.09} & \textbf{7.39} & \textbf{8.25} \\
 & Nano-Banana-2 & 8.38 & 8.39 & 8.13 & 7.81 & 7.97 & 8.93 & 7.81 & 7.72 & 6.97 & 7.66 & 6.97 & 7.89 \\
 & Qwen-Image-2 & 7.18 & 7.01 & 6.90 & 6.72 & 6.83 & 7.59 & 6.29 & 6.32 & 6.34 & 6.97 & 5.88 & 6.73 \\
\midrule
\multirow{8}{*}{Open-source} & Qwen-Image & 6.50 & 7.09 & 6.70 & 5.85 & 6.04 & 7.04 & 5.34 & 4.84 & 5.38 & 5.92 & 4.55 & 5.93 \\
 & Z-Image-Turbo & 6.66 & 7.72 & 7.16 & 6.51 & 6.45 & 7.49 & 5.68 & 5.25 & 6.18 & 6.73 & 4.99 & 6.44 \\
 & LongCat-Image & 4.69 & 6.91 & 6.16 & 4.85 & 5.05 & 5.88 & 3.97 & 3.56 & 4.61 & 5.26 & 3.57 & 4.96 \\
 & FLUX.2-dev & 6.60 & 8.24 & 7.56 & 6.46 & 6.77 & 7.74 & 5.70 & 4.20 & 5.93 & 6.65 & 4.38 & 6.38 \\
 & HiDream-O1-Image & 6.82 & 5.97 & 5.83 & 5.25 & 5.64 & 6.20 & 4.85 & 4.50 & 4.81 & 5.32 & 4.09 & 5.39 \\
 & BAGEL & 3.50 & 7.50 & 6.64 & 4.41 & 5.68 & 5.83 & 4.03 & 2.52 & 4.67 & 5.24 & 2.74 & 4.80 \\
 & OmniGen2 & 3.89 & 6.83 & 6.01 & 4.19 & 4.80 & 5.31 & 3.65 & 2.38 & 3.75 & 4.03 & 2.41 & 4.30 \\
 & \textbf{InfoVis (ours)} & \textbf{8.42} & \textbf{8.32} & \textbf{7.98} & \textbf{7.62} & \textbf{7.81} & \textbf{8.78} & \textbf{7.53} & \textbf{7.34} & \textbf{6.71} & \textbf{7.42} & \textbf{6.60} & \textbf{7.68} \\
\bottomrule
\end{tabular}%
}
\endgroup
\end{table}

\begin{table}[!htbp]
\centering
\caption{Detailed template-conditioned generation results. Dimensions follow Table~\ref{tab:t2i_fine_grained_results}; Overall averages the same eleven dimensions. Bold marks the best score within each model group. Higher is better.}
\label{tab:template_fine_grained_results}
\begingroup
\scriptsize
\setlength{\tabcolsep}{4.5pt}
\renewcommand{\arraystretch}{1.00}
\resizebox{\linewidth}{!}{%
\begin{tabular}{@{}ll*{12}{r}@{}}
\toprule
\multirow{2}{*}{Type} & \multirow{2}{*}{Model} & \multicolumn{1}{c}{CF} & \multicolumn{6}{c}{IF} & \multicolumn{4}{c}{AQ} & \multirow{2}{*}{Overall} \\
\cmidrule(lr){3-3}\cmidrule(lr){4-9}\cmidrule(lr){10-13}
 & & Text & BG & Color & Typo & Layout & Mood & Block & Clarity & LayoutQ & StyleQ & Prof & \\
\midrule
\multirow{3}{*}{Proprietary} & GPT-Image-2 & 8.19 & \textbf{9.08} & \textbf{8.70} & \textbf{8.50} & \textbf{7.51} & \textbf{9.34} & \textbf{7.93} & \textbf{7.93} & \textbf{7.22} & \textbf{8.02} & \textbf{7.32} & \textbf{8.16} \\
 & Nano-Banana-2 & 7.59 & 8.85 & 8.26 & 7.97 & 6.74 & 8.90 & 6.96 & 7.38 & 6.73 & 7.47 & 6.74 & 7.60 \\
 & Qwen-Image-2 & \textbf{8.23} & 6.65 & 6.49 & 6.07 & 5.89 & 6.93 & 5.68 & 5.85 & 5.53 & 6.24 & 5.35 & 6.26 \\
\midrule
\multirow{7}{*}{Open-source} & Qwen-Image-Edit & 3.10 & 7.25 & 7.02 & 6.51 & 4.46 & 6.63 & 3.98 & 5.19 & 5.49 & 6.11 & 4.83 & 5.51 \\
 & LongCat-Image-Edit & 4.86 & 7.08 & 6.51 & 5.69 & 5.46 & 6.70 & 4.52 & 4.51 & 5.37 & 5.88 & 4.41 & 5.54 \\
 & FLUX.2-dev & 6.28 & 8.84 & 7.96 & 6.85 & 6.57 & 8.27 & 5.77 & 4.12 & 5.89 & 6.56 & 4.30 & 6.49 \\
 & HiDream-O1-Image & 5.97 & 4.96 & 5.44 & 5.08 & 5.53 & 5.54 & 4.30 & 4.26 & 4.71 & 5.04 & 3.97 & 4.98 \\
 & BAGEL & 3.31 & 6.45 & 6.04 & 3.96 & 5.03 & 5.62 & 3.90 & 2.65 & 4.44 & 4.98 & 2.85 & 4.48 \\
 & OmniGen2 & 3.75 & 7.97 & 6.98 & 4.52 & 5.15 & 6.28 & 4.04 & 2.48 & 3.97 & 4.31 & 2.59 & 4.73 \\
 & \textbf{InfoVis (ours)} & \textbf{8.14} & \textbf{8.85} & \textbf{8.24} & \textbf{7.87} & \textbf{7.29} & \textbf{8.94} & \textbf{7.33} & \textbf{7.25} & \textbf{6.72} & \textbf{7.41} & \textbf{6.62} & \textbf{7.70} \\
\bottomrule
\end{tabular}%
}
\endgroup
\end{table}

\begin{table}[!htbp]
\centering
\caption{Detailed multi-image grounded generation results. In addition to Text Rendering, CF includes Multi-image Consistency (MIC), which measures preservation and use of supplied assets. Other dimensions follow Table~\ref{tab:t2i_fine_grained_results}. Overall averages all twelve dimensions. Bold marks the best score within each model group. Higher is better.}
\label{tab:multi_image_fine_grained_results}
\begingroup
\scriptsize
\setlength{\tabcolsep}{4.5pt}
\renewcommand{\arraystretch}{1.00}
\resizebox{\linewidth}{!}{%
\begin{tabular}{@{}ll*{13}{r}@{}}
\toprule
\multirow{2}{*}{Type} & \multirow{2}{*}{Model} & \multicolumn{2}{c}{CF} & \multicolumn{6}{c}{IF} & \multicolumn{4}{c}{AQ} & \multirow{2}{*}{Overall} \\
\cmidrule(lr){3-4}\cmidrule(lr){5-10}\cmidrule(lr){11-14}
 & & Text & MIC & BG & Color & Typo & Layout & Mood & Block & Clarity & LayoutQ & StyleQ & Prof & \\
\midrule
\multirow{3}{*}{Proprietary} & GPT-Image-2 & \textbf{8.38} & \textbf{9.47} & \textbf{8.74} & \textbf{8.29} & \textbf{8.26} & \textbf{8.20} & \textbf{9.16} & \textbf{8.00} & \textbf{7.83} & \textbf{7.24} & \textbf{8.06} & \textbf{7.30} & \textbf{8.24} \\
 & Nano-Banana-2 & 7.74 & 9.46 & 8.37 & 7.87 & 7.65 & 7.52 & 8.78 & 7.39 & 7.34 & 6.77 & 7.66 & 6.79 & 7.78 \\
 & Qwen-Image-2 & 7.25 & 6.69 & 6.42 & 6.31 & 6.23 & 5.97 & 6.81 & 5.75 & 5.50 & 5.44 & 6.18 & 5.07 & 6.14 \\
\midrule
\multirow{7}{*}{Open-source} & Qwen-Image-Edit & 3.74 & 4.94 & 5.96 & 5.53 & 3.96 & 4.20 & 5.37 & 3.40 & 3.41 & 4.41 & 5.27 & 3.39 & 4.47 \\
 & LongCat-Image-Edit & 4.58 & 5.91 & 5.97 & 5.79 & 4.60 & 4.59 & 5.76 & 3.91 & 4.07 & 4.67 & 5.43 & 3.88 & 4.93 \\
 & FLUX.2-dev & 6.48 & 6.60 & 8.44 & 7.54 & 6.54 & 6.72 & 7.84 & 5.69 & 4.29 & 5.76 & 6.40 & 4.34 & 6.39 \\
 & HiDream-O1-Image & 5.79 & 5.41 & 5.32 & 5.17 & 4.72 & 4.76 & 5.38 & 4.05 & 3.90 & 4.38 & 4.76 & 3.64 & 4.77 \\
 & BAGEL & 3.22 & 4.26 & 5.85 & 5.35 & 3.80 & 4.74 & 5.07 & 3.75 & 2.67 & 4.09 & 4.64 & 2.81 & 4.19 \\
 & OmniGen2 & 3.67 & 4.45 & 6.35 & 5.71 & 3.89 & 4.59 & 5.27 & 3.60 & 2.47 & 3.74 & 4.16 & 2.53 & 4.20 \\
 & \textbf{InfoVis (ours)} & \textbf{8.22} & \textbf{9.20} & \textbf{8.45} & \textbf{7.87} & \textbf{7.74} & \textbf{7.70} & \textbf{8.80} & \textbf{7.49} & \textbf{7.21} & \textbf{6.78} & \textbf{7.57} & \textbf{6.67} & \textbf{7.81} \\
\bottomrule
\end{tabular}%
}
\endgroup
\end{table}

\begin{table}[t]
\centering
\caption{Results for all nine editing operations. Text, Typo, Image, Theme, and BG denote text modification, typography changes, image replacement, theme switching, and background updates. Each score is the geometric mean of Semantic Consistency (SC) and Presentation Quality (PQ); Overall uses overall SC and PQ. Bold marks the best score within each model group. Higher is better.}
\label{tab:edit_fine_grained_results}
\begingroup
\scriptsize
\setlength{\tabcolsep}{4.5pt}
\renewcommand{\arraystretch}{1.00}
\resizebox{\linewidth}{!}{%
\begin{tabular}{@{}ll*{10}{r}@{}}
\toprule
Type & Model & Text & Color & Font & Typo & Remove & Add & Image & Theme & BG & Overall \\
\midrule
\multirow{3}{*}{Proprietary} & GPT-Image-2 & \textbf{9.03} & \textbf{8.83} & \textbf{8.52} & \textbf{8.64} & \textbf{8.42} & \textbf{8.98} & \textbf{9.44} & 8.77 & \textbf{8.74} & \textbf{8.88} \\
 & Nano-Banana-2 & 8.93 & 8.80 & 8.30 & 8.53 & 8.07 & 8.85 & 9.33 & \textbf{9.26} & 8.62 & 8.83 \\
 & Qwen-Image-2 & 5.82 & 5.67 & 3.84 & 4.54 & 2.80 & 6.15 & 6.43 & 7.75 & 5.79 & 5.66 \\
\midrule
\multirow{7}{*}{Open-source} & Qwen-Image-Edit & 7.02 & 6.26 & 5.95 & 6.80 & 7.66 & 7.95 & 8.74 & \textbf{8.13} & 7.84 & 7.57 \\
 & LongCat-Image-Edit & 5.52 & 6.58 & 4.06 & 5.22 & 7.23 & 6.66 & 7.95 & 7.41 & 6.94 & 6.63 \\
 & FLUX.2-dev & 6.04 & 5.99 & 4.42 & 4.68 & 5.84 & 5.91 & 8.22 & 7.88 & \textbf{8.25} & 6.63 \\
 & HiDream-O1-Image & 2.16 & 1.34 & 1.34 & 2.92 & 0.74 & 2.36 & 3.94 & 5.75 & 1.40 & 2.90 \\
 & BAGEL & 3.31 & 5.74 & 3.81 & 4.99 & 4.13 & 5.04 & 7.27 & 6.72 & 6.49 & 5.61 \\
 & OmniGen2 & 1.83 & 3.59 & 3.83 & 2.00 & 3.04 & 3.26 & 4.69 & 5.83 & 3.62 & 3.84 \\
 & \textbf{InfoVis (ours)} & \textbf{8.99} & \textbf{8.72} & \textbf{8.24} & \textbf{8.46} & \textbf{8.09} & \textbf{8.34} & \textbf{8.79} & 7.94 & 8.00 & \textbf{8.41} \\
\bottomrule
\end{tabular}%
}
\endgroup
\end{table}

\subsection{Source-Stratified Evaluation}
\label{app:source_split_results}

VisSlideBench combines records derived from real presentation decks with
synthetic records generated during benchmark construction: 167 real and 150
synthetic text-to-slide records, 147 and 150 template-conditioned records, and
179 and 150 multi-image records. We recompute the three generation-task scores
on each partition. For each slide, the score is the arithmetic mean of CF, IF,
and AQ; we average within each task and source, and the macro score weights
the three task means equally.

\begin{table*}[t]
\centering
\caption{Generation performance by benchmark source. R and S denote real and
synthetic records. Each task score averages CF, IF, and AQ over the partition;
Macro averages the three task scores. $|\Delta|$ is the absolute
Real--Synthetic macro gap. Bold marks the best score within each model group.
Higher is better except for $|\Delta|$.}
\label{tab:source_split_results}
\begingroup
\scriptsize
\setlength{\tabcolsep}{3.4pt}
\renewcommand{\arraystretch}{1.04}
\resizebox{\textwidth}{!}{%
\begin{tabular}{@{}ll*{9}{r}@{}}
\toprule
\multirow{2}{*}{Type} & \multirow{2}{*}{Model}
& \multicolumn{2}{c}{T2I}
& \multicolumn{2}{c}{Template}
& \multicolumn{2}{c}{Multi-image}
& \multicolumn{2}{c}{Macro}
& \multirow{2}{*}{$|\Delta|$} \\
\cmidrule(lr){3-4}\cmidrule(lr){5-6}\cmidrule(lr){7-8}\cmidrule(lr){9-10}
& & R & S & R & S & R & S & R & S & \\
\midrule
\multirow{3}{*}{Proprietary}
& GPT-Image-2       & \textbf{8.26} & \textbf{8.28} & \textbf{8.08} & \textbf{8.13} & \textbf{8.36} & \textbf{8.28} & \textbf{8.23} & \textbf{8.23} & 0.004 \\
& Nano-Banana-2     & 8.00 & 7.92 & 7.37 & 7.70 & 7.95 & 7.81 & 7.77 & 7.81 & 0.039 \\
& Qwen-Image-2      & 6.67 & 6.98 & 7.15 & 6.36 & 5.98 & 6.60 & 6.60 & 6.65 & 0.046 \\
\midrule
\multirow{6}{*}{Open-source}
& FLUX.2-dev        & 6.25 & 6.41 & 6.41 & 6.17 & 6.15 & 6.46 & 6.27 & 6.35 & 0.076 \\
& FLUX.2-klein-base-9B
                    & 6.15 & 6.25 & 6.53 & 6.20 & 6.15 & 6.47 & 6.28 & 6.31 & 0.029 \\
& HiDream-O1-Image  & 5.81 & 5.60 & 5.44 & 4.99 & 4.61 & 5.22 & 5.28 & 5.27 & 0.017 \\
& BAGEL             & 4.18 & 4.49 & 4.03 & 4.10 & 3.81 & 4.26 & 4.01 & 4.28 & 0.273 \\
& OmniGen2          & 4.03 & 4.08 & 4.37 & 4.23 & 3.92 & 4.23 & 4.11 & 4.18 & 0.068 \\
& \textbf{InfoVis (ours)}
                    & \textbf{7.76} & \textbf{7.88} & \textbf{7.83} & \textbf{7.65} & \textbf{7.98} & \textbf{7.86} & \textbf{7.86} & \textbf{7.80} & 0.060 \\
\bottomrule
\end{tabular}%
}
\endgroup
\end{table*}

Table~\ref{tab:source_split_results} shows that the ranking is stable across
sources. Across the nine models, the Real--Synthetic Spearman rank
correlations are 1.000 for T2I, 0.983 for template-conditioned generation,
0.929 for multi-image generation, and 0.950 for the macro score. InfoVis
ranks first among open-source models for every task on both sources. Its Real
and Synthetic macro scores are 7.86 and 7.80, a gap of 0.060 against a mean
gap of 0.068 across models. The two partitions contain different slide
content, so the table shows robustness of the ranking to the source, not
independence from it.

\subsection{Multi-Image SFT by Benchmark Source}
\label{app:multi_image_sft_split}

The multi-image SFT pool is fully synthetic: its targets are rendered by
Nano-Banana-2 and its conditioning assets are regions cropped from those
renders (Appendix~\ref{app:data_sft}). The multi-image split of VisSlideBench
instead contains 179 cases from real decks and 150 synthetic cases
(Appendix~\ref{app:benchmark_composition}).
Table~\ref{tab:multi_image_sft_split} reports CF, IF, and AQ on the two
sources for the checkpoints of Table~\ref{tab:ablation_training_stages}.

\begin{table}[!htbp]
\centering
\caption{Multi-image grounded generation by benchmark source for the
training-stage checkpoints. R: cases from real decks (179); S: synthetic cases
(150). Each entry averages the per-case score over the cases with valid CF,
IF, and AQ. Bold marks the best score in each column.}
\label{tab:multi_image_sft_split}
\begingroup
\scriptsize
\setlength{\tabcolsep}{5pt}
\renewcommand{\arraystretch}{1.02}
\begin{tabular}{@{}l*{6}{r}@{}}
\toprule
\multirow{2}{*}{Model}
& \multicolumn{3}{c}{Real (R)} & \multicolumn{3}{c}{Synthetic (S)} \\
\cmidrule(lr){2-4}\cmidrule(lr){5-7}
& CF & IF & AQ & CF & IF & AQ \\
\midrule
Stage-2 CPT: four-task mixture & 8.60 & 7.90 & 6.34 & 8.12 & 7.11 & 6.21 \\
Shared SFT                     & 8.64 & 8.05 & 6.68 & 8.34 & 7.53 & 6.97 \\
Shared SFT + T2I SFT           & 8.50 & 8.12 & 6.84 & 7.92 & 7.65 & 7.05 \\
Shared SFT + Template SFT      & 7.62 & 8.08 & 6.68 & 7.36 & 7.61 & 6.95 \\
Shared SFT + Multi-image SFT   & 7.99 & 7.55 & 5.94 & 8.29 & 7.23 & 6.67 \\
Shared SFT + Edit SFT          & 8.43 & 8.01 & 6.44 & 7.91 & 7.56 & 6.79 \\
\midrule
\textbf{InfoVis}               & \textbf{8.79} & \textbf{8.27} & \textbf{6.90} & \textbf{8.61} & \textbf{7.70} & \textbf{7.26} \\
\bottomrule
\end{tabular}
\endgroup
\end{table}

Multi-image SFT lowers all three metrics on real cases (CF 8.64 to 7.99, IF
8.05 to 7.55, AQ 6.68 to 5.94), whereas synthetic cases lose much less (CF
8.34 to 8.29, IF 7.53 to 7.23, AQ 6.97 to 6.67) and the asset-preservation component
of CF rises from 8.03 to 8.29. The typical failure on real cases is an asset that
is pasted too large or redrawn, displacing or overlapping the surrounding text
(Appendix~\ref{app:failure_cases}). Template SFT, whose pool is also
synthetic, lowers multi-image CF on both sources (7.62 and 7.36). Fusing the
task checkpoints with those that retain the real-slide priors of continual
pretraining recovers both sources, and the fused model is the best in every
column.

\subsection{Prompt-Length Truncation}
\label{app:truncation_audit}

Short requests run 62 to 376 tokens, detailed descriptions 253 to 828, and
StyleJSON specifications 1,102 to 3,731, so a 1,024-token window clips every
StyleJSON prompt and nothing else. Token counts use the Qwen3 tokenizer,
except for LongCat-Image, which is counted under its own tokenization.
Table~\ref{tab:truncation_audit} lists the
baselines whose runs used a window shorter than the prompts, with the window
taken from the run behind Table~\ref{tab:main_slidebench_results}.
LongCat-Image fixes a 512-token window and tokenizes quoted spans word by
word, so it also clips 40\% of the short and 90\% of the long prompts.
Qwen-Image-2 is served through an API that does not publish its window; text
placed after token 1,024 of a StyleJSON prompt almost never appears in its
output, while text before that point does, so we treat 1,024 as its effective window. For each
baseline we rescore InfoVis and the baseline on the prompts the baseline
received in full. Every gap narrows. InfoVis stays ahead of the three
open-source baselines on all three metrics and ties Qwen-Image-2 on
text-to-slide generation. On the 80 prompts that no model truncates, InfoVis
remains the best open-source model behind GPT-Image-2 and Nano-Banana-2.
Truncation does not explain the StyleJSON pattern in
Figure~\ref{fig:scaling_prompt_categories}b: FLUX.2-dev,
FLUX.2-klein-base-9B, BAGEL, and OmniGen2 receive every prompt in full and
score 4.71, 3.43, 2.30, and 1.81 in StyleJSON Content Faithfulness, against
7.95 for InfoVis.

\begin{table}[!htbp]
\centering
\caption{Baselines whose text window clips VisSlideBench prompts. Clipped
counts are per category (115 short, 109 long, 93 StyleJSON). $\Delta$ is
InfoVis minus the baseline in CF/IF/AQ, on the prompts the baseline received
in full and on the full T2I split of Table~\ref{tab:main_slidebench_results}.}
\label{tab:truncation_audit}
\begingroup
\scriptsize
\setlength{\tabcolsep}{5pt}
\begin{tabular}{@{}lrrrrll@{}}
\toprule
Baseline & Window & Clipped (S/L/J) & Full prompts & $\Delta$ on full prompts & $\Delta$ in Table~\ref{tab:main_slidebench_results} \\
\midrule
Qwen-Image    & 1,024 & 0 / 0 / 93   & 224 & $+0.56$ / $+0.96$ / $+1.22$ & $+1.92$ / $+1.67$ / $+1.85$ \\
Z-Image-Turbo & 1,024 & 0 / 0 / 93   & 224 & $+0.56$ / $+0.46$ / $+0.89$ & $+1.76$ / $+1.17$ / $+1.23$ \\
LongCat-Image & 512   & 46 / 98 / 93 & 80  & $+2.05$ / $+1.54$ / $+2.27$ & $+3.73$ / $+2.54$ / $+2.77$ \\
Qwen-Image-2  & 1,024 & 0 / 0 / 93   & 224 & $-0.05$ / $+0.35$ / $-0.30$ & $+1.24$ / $+1.12$ / $+0.64$ \\
\bottomrule
\end{tabular}
\endgroup
\end{table}

\section{Document-to-Slide Workflow}
\label{app:workflow}

\begin{figure}[htbp]
    \centering
    \includegraphics[width=\linewidth]{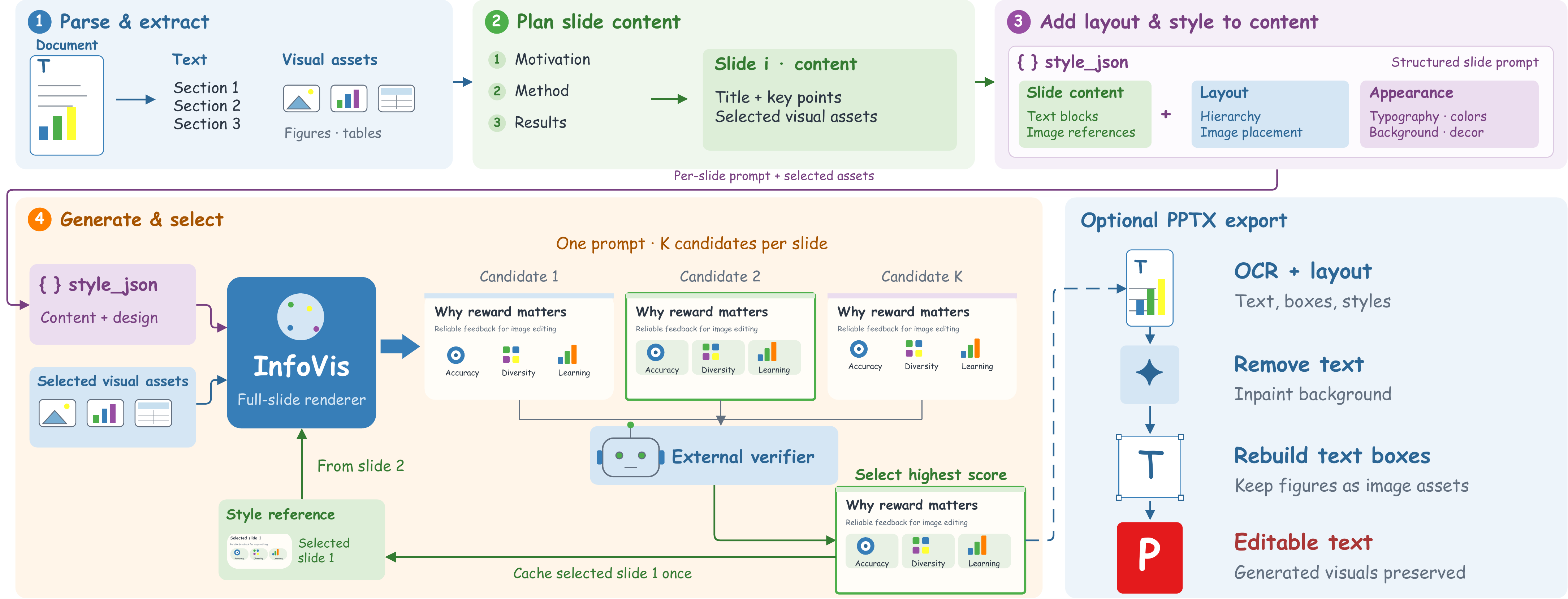}
    \caption{Document-to-slide workflow with InfoVis. Content planning and
    design planning produce structured per-slide prompts. InfoVis generates
    candidates that an external verifier ranks. The selected first slide
    serves as a style reference for subsequent slides. Optional PowerPoint
    export reconstructs editable text while retaining visual assets as images.
    Candidate slides are schematic illustrations.}
    \label{fig:document_to_slide_workflow}
\end{figure}

The workflow uses InfoVis to render complete slide images from an input
document. It connects document parsing, content planning, design planning,
candidate selection, and optional PowerPoint export, as illustrated in
Figure~\ref{fig:document_to_slide_workflow}.

\paragraph{Document parsing and content planning.}
A parser extracts the document's text and visual assets, including figures
and tables. A content planner (GPT-4.1~\citep{openai2025gpt41}) uses these
materials to form an outline for the presentation and a content plan for each
slide. Each plan specifies the text
and visual assets to include on that page. The design planner then determines
how to arrange and style this content.

\paragraph{Structured design planning.}
The design planner (also GPT-4.1) converts each content plan into a \texttt{style\_json}
prompt that specifies text blocks, layout hierarchy, image placement,
typography, colors, background, and decorative elements. InfoVis renders
each prompt together with the selected visual assets.

\paragraph{Candidate generation and selection.}
For each slide $i$, InfoVis generates $K$ candidates from its design prompt
$z_i$ and input images $\mathcal{I}_i$. An external verifier $V$
(Qwen3.5-27B~\citep{qwen3.5}) scores each candidate against the prompt, and
the workflow keeps the highest-scoring candidate
(Appendix~\ref{app:experimental_setup} gives the verifier configuration and $K$):
\begin{equation}
    y_i^{(k)} = G_\theta(z_i, \mathcal{I}_i; \epsilon_k),
    \qquad
    k_i^* = \arg\max_{k\in\{1,\ldots,K\}} V(y_i^{(k)}, z_i),
    \qquad
    y_i^* = y_i^{(k_i^*)}.
\end{equation}
Here, $G_\theta$ is InfoVis and $\epsilon_k$ is the sampling randomness. The
first slide is generated without a style reference; its selected candidate
$y_1^*$ then serves as the reference image for every later slide, which keeps
the visual style consistent while each slide follows its own content plan. The SlidesGen-Bench run in Table~\ref{tab:aesthetics_eval} uses $K=2$.

\paragraph{PowerPoint export.}
The selected slide images can optionally be converted into PowerPoint files
with editable text. OCR and layout analysis recover the text content,
bounding boxes, and basic styles. The workflow removes the detected text and
inpaints those regions to create a background image, then reconstructs the
recognized text as native PowerPoint text boxes. Figures and tables are
retained as cropped image assets at their detected positions, so charts and
tables remain images rather than native, editable objects.

\section{Limitations}
\label{app:limitations}

VisSlideBench-Edit is held out from the same construction pipeline as the
editing training data, so it measures in-distribution editing quality;
out-of-distribution evidence comes from BizGenEval for generation and from
SlidesGen-Bench for the workflow. Aesthetic Quality remains the dimension
with the largest gap to proprietary models, and knowledge- and
reasoning-intensive slides, as in the BizGenEval hard split, lie outside what
slide-specific supervision targets. Our generation SFT targets and the synthetic half of
VisSlideBench-Gen are produced with a proprietary model, which ties both the
training distribution and those evaluation cases to its style;
Table~\ref{tab:source_split_results} therefore reports generation results
separately for real and synthetic cases. Finally, the
data and benchmark cover English-language slides only.
Appendix~\ref{app:failure_cases} illustrates failures in visual asset fidelity
and the preservation of content outside an editing target.

\clearpage
\section{Qualitative Results}
\label{app:qualitative_results}

\begin{figure}[H]
    \centering
    \vspace{20pt}    
    \includegraphics[width=\linewidth,height=0.76\textheight,keepaspectratio]{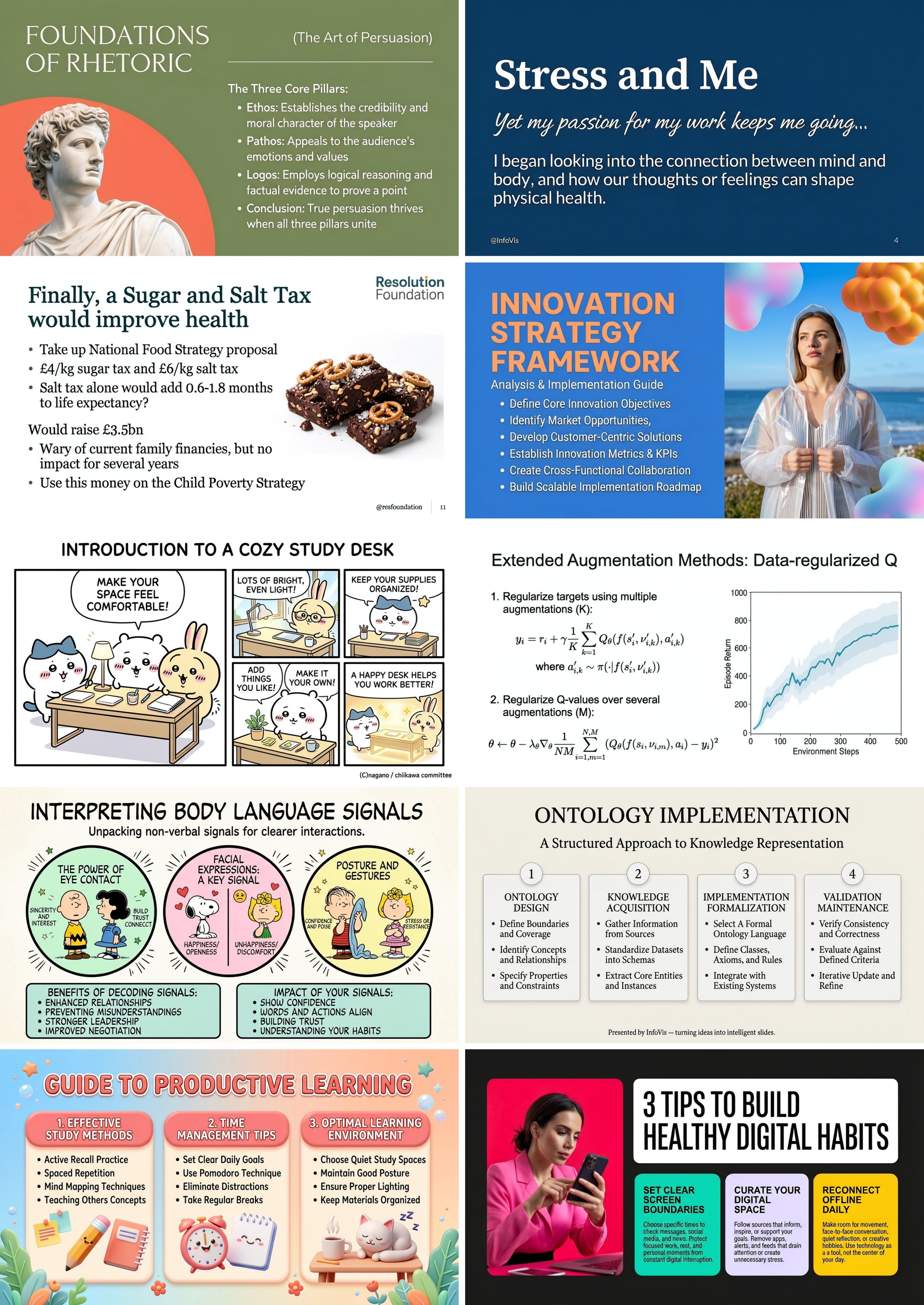}
    \caption{Showcases of InfoVis in text-to-slide image generation.}
    \label{fig:t2i_case}
\end{figure}

\begin{figure}[H]
    \centering
    \vspace{20pt}    
    \includegraphics[width=\linewidth,height=0.88\textheight,keepaspectratio]{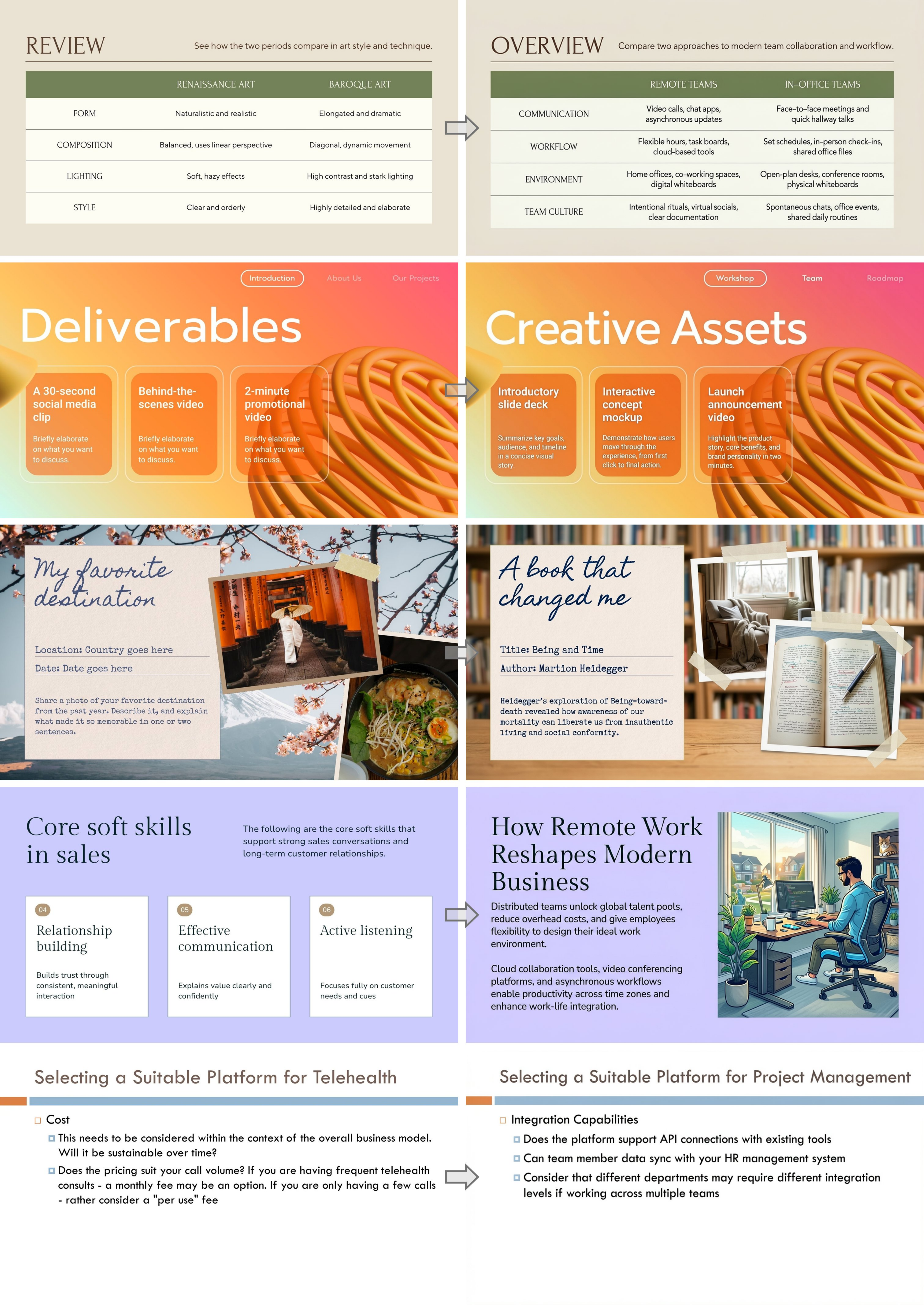}
    \caption{Showcases of InfoVis in template-conditioned slide image generation.}
    \label{fig:ref_case}
\end{figure}

\begin{figure}[H]
    \centering
    \vspace{20pt}    
    \includegraphics[width=\linewidth,height=0.88\textheight,keepaspectratio]{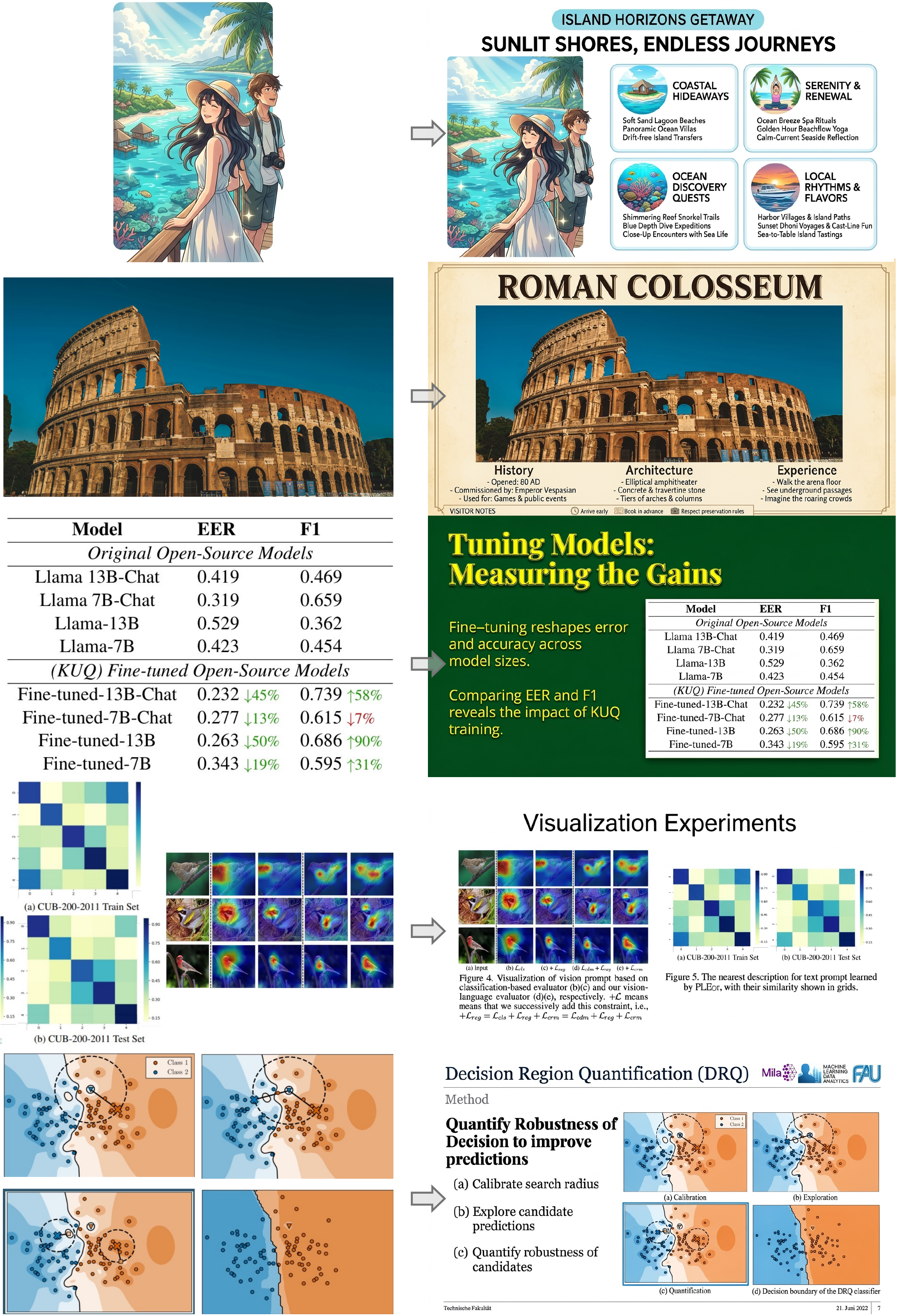}
    \caption{Showcases of InfoVis in multi-image grounded slide image generation.}
    \label{fig:ic_case}
\end{figure}

\begin{figure}[H]
    \centering
    \vspace{20pt}    
    \includegraphics[width=\linewidth,height=0.88\textheight,keepaspectratio]{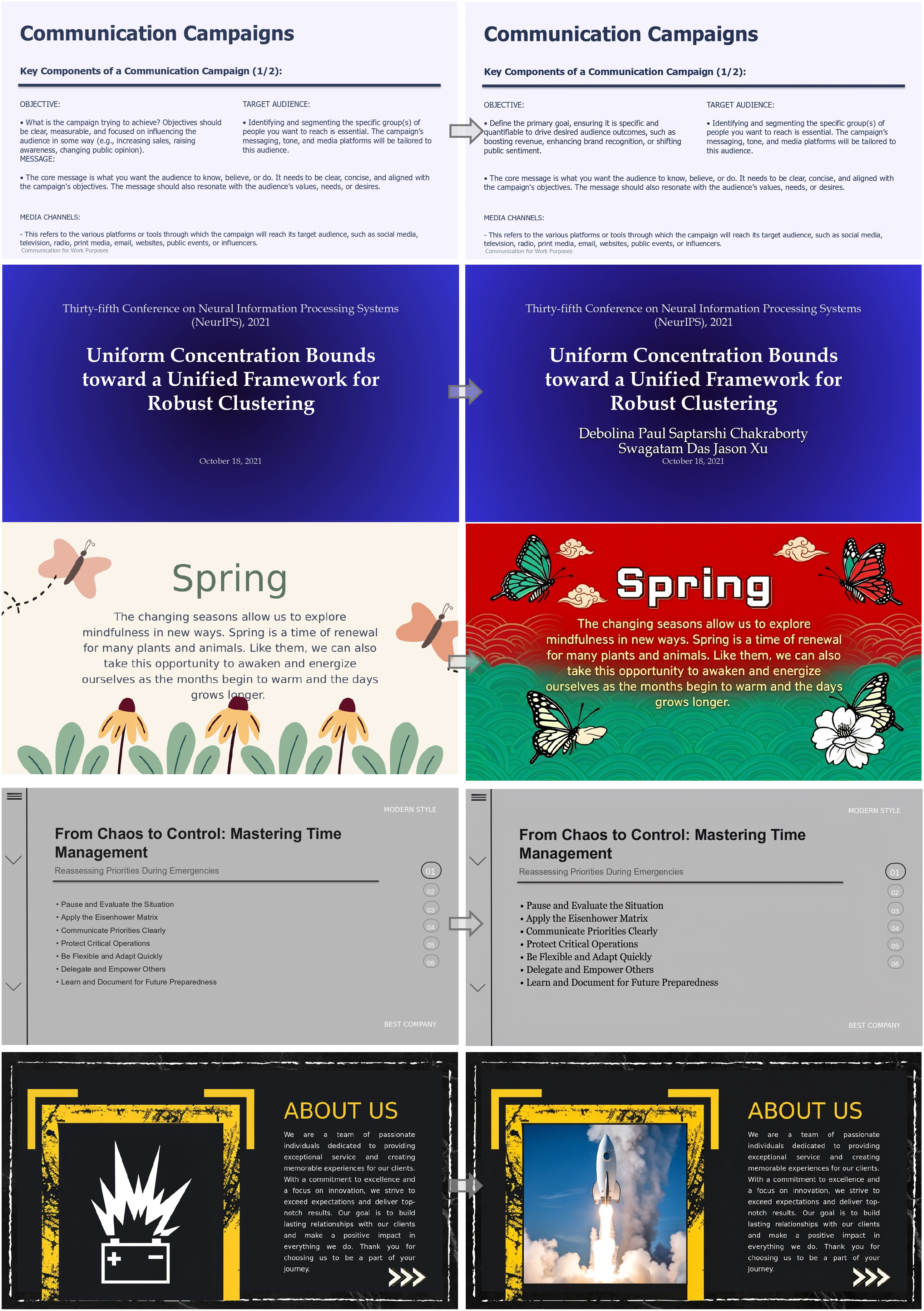}
    \caption{Showcases of InfoVis in  slide image editing.}
    \label{fig:edit_case}
\end{figure}

\begin{figure}[H]
    \centering
    \vspace{20pt}    
    \includegraphics[width=\linewidth,height=0.88\textheight,keepaspectratio]{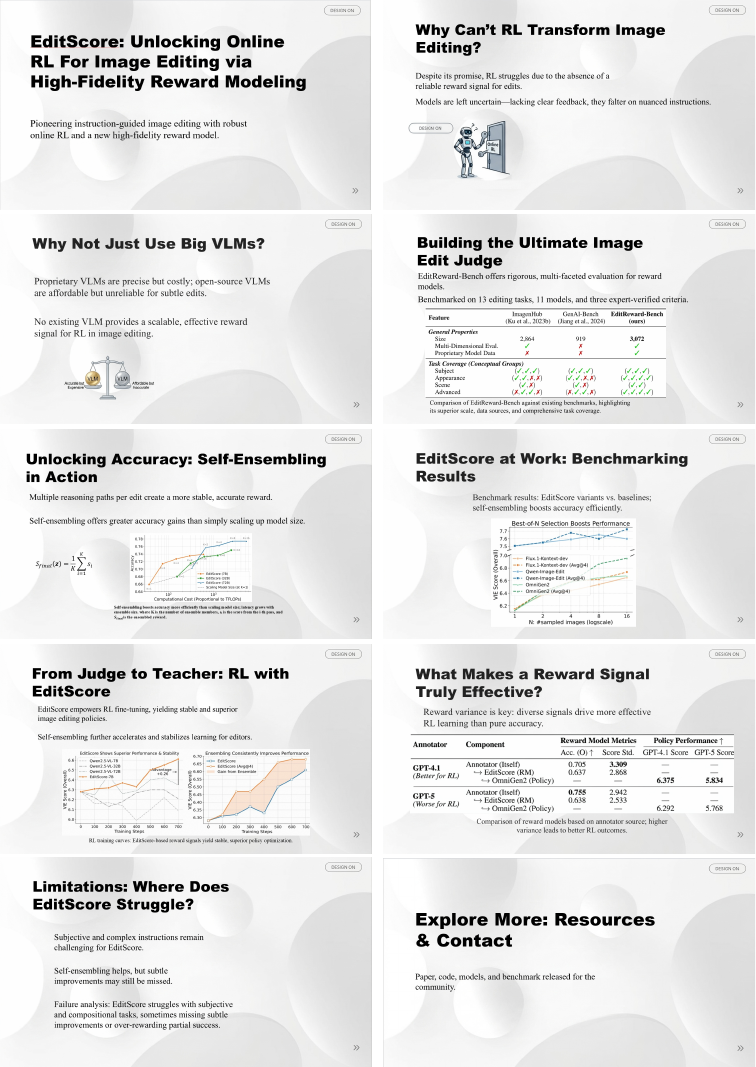}
    \caption{Showcases of the document-to-slide workflow built on InfoVis.}
    \label{fig:workflow}
\end{figure}

Figures~\ref{fig:t2i_case}, \ref{fig:ref_case}, \ref{fig:ic_case},
and~\ref{fig:edit_case} show qualitative results across the four slide tasks.
InfoVis generates coherent slide layouts from text, follows reference styles,
faithfully integrates multiple visual assets, and performs both local and
global edits while preserving unrelated content. Figure~\ref{fig:workflow}
further shows the document-to-slide workflow producing visually consistent
multi-slide presentations with relevant textual and visual content.

\clearpage

\clearpage

\clearpage

\clearpage

\clearpage

\section{Failure Cases}
\label{app:failure_cases}

\begin{figure}[H]
    \centering
    \includegraphics[width=\linewidth]{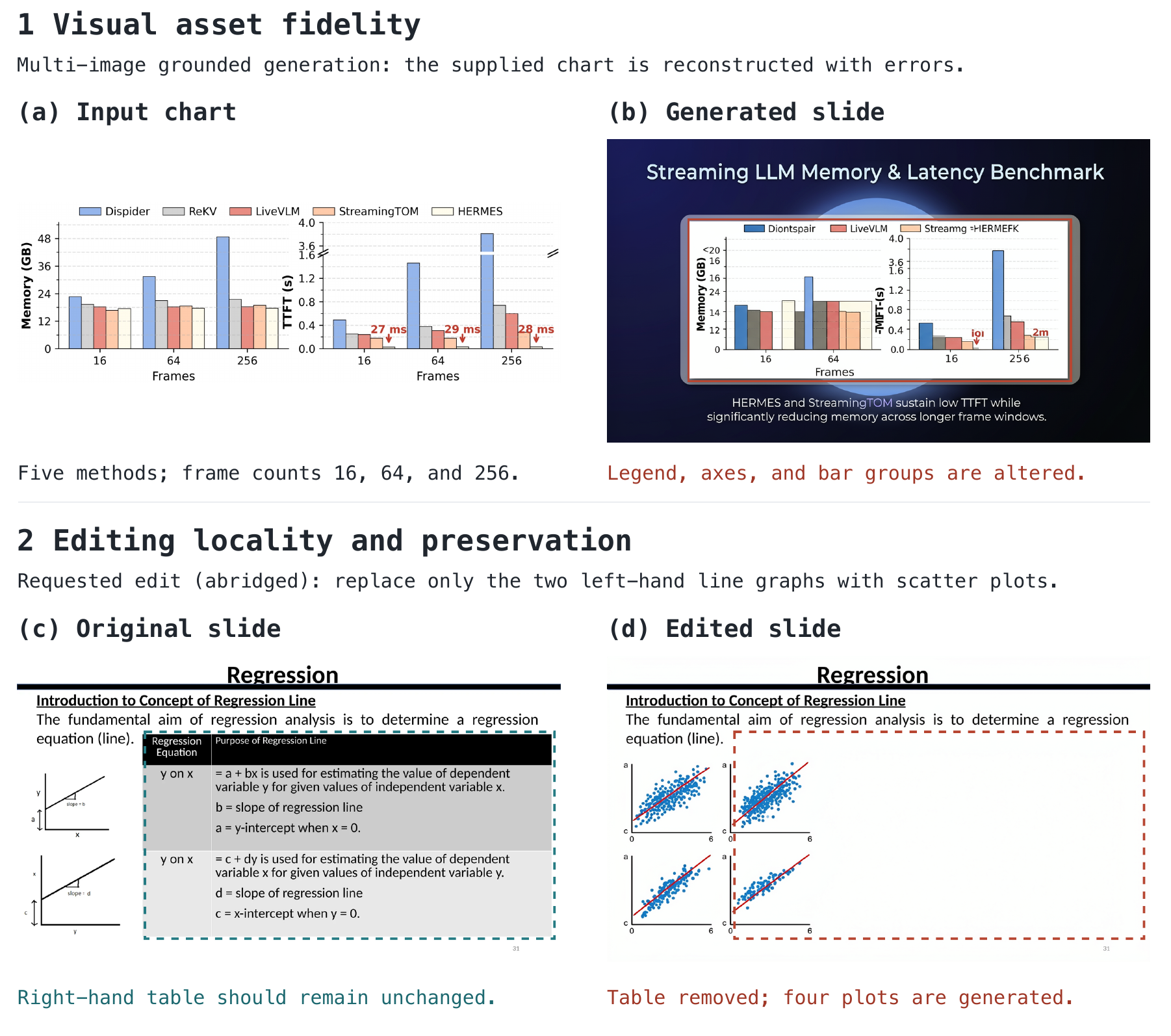}
    \caption{\textbf{Failure cases of InfoVis.}
    (a)--(b) A supplied chart and the generated slide show errors in visual
    asset fidelity; the solid outline marks the reconstructed chart.
    (c)--(d) An original slide and its edited output show changes beyond
    the requested left-hand graphs. Dashed outlines mark the table region
    that should have been preserved. The editing instruction above the
    panels is abridged.}
    \label{fig:failure_cases}
\end{figure}

Figure~\ref{fig:failure_cases} shows two failure cases of InfoVis, one in
content fidelity and one in editing locality.

\paragraph{Visual asset fidelity.}
In multi-image grounded generation, a supplied chart can be reconstructed
rather than faithfully preserved. In panels (a)--(b), the generated slide
retains the broad appearance of the chart but corrupts legend entries and
axis labels and changes the bar-group structure.

\paragraph{Editing locality and preservation.}
In panels (c)--(d), the instruction asks to replace only the two line graphs
on the left with vertically stacked scatter plots. The output instead
contains four plots and removes the table on the right, which lies outside
the requested editing target. The model identifies what to change but does
not reliably confine the change to the requested region.

\end{document}